\documentclass[10pt,twocolumn,letterpaper]{article}

\usepackage{iccv}
\usepackage{times}
\usepackage{epsfig}
\usepackage{graphicx}
\usepackage{subcaption}
\usepackage{amsmath}
\usepackage{amssymb}
\usepackage{stmaryrd}
\usepackage{flushend}
\usepackage{enumitem}
\usepackage{booktabs}
\usepackage{multirow}

\usepackage[pagebackref=true,breaklinks=true,colorlinks,citecolor=NavyBlue,bookmarks=false]{hyperref}

\usepackage{bm}
\usepackage[dvipsnames]{xcolor}
\usepackage{soul,colortbl}

\definecolor{ggreen}{RGB}{15,157,88}
\definecolor{gred}{RGB}{219,68,55}
\definecolor{mmm}{RGB}{219,38,95}

\definecolor{flat}{RGB}{128,64,128}
\definecolor{construction}{RGB}{70,70,70}
\definecolor{object}{RGB}{220,220,0}
\definecolor{nature}{RGB}{107,142,35}
\definecolor{sky}{RGB}{70,130,180}
\definecolor{human}{RGB}{220,20,60}
\definecolor{vehicle}{RGB}{0,0,142}

\newcommand{\Flat}{\textcolor{flat}{flat}}
\newcommand{\construction}{\textcolor{construction}{construction}}
\newcommand{\object}{\textcolor{object}{object}}
\newcommand{\nature}{\textcolor{nature}{nature}}
\newcommand{\sky}{\textcolor{sky}{sky}}
\newcommand{\human}{\textcolor{human}{human}}
\newcommand{\vehicle}{\textcolor{vehicle}{vehicle}}

\newcommand{\up}[1]{\textcolor{ggreen}{\uparrow{#1}}}
\newcommand{\down}[1]{\textcolor{gred}{\downarrow{#1}}}

\newcolumntype{C}[1]{>{\centering\arraybackslash}p{#1}}	
\newcolumntype{L}[1]{>{\raggedright\arraybackslash}p{#1}}
\newcolumntype{R}[1]{>{\raggedleft\arraybackslash}p{#1}}

\newcommand{\mv}[1]{\ensuremath{\bm{#1}}} 
\newcommand{\bx}{\ensuremath{\mathbf{x}}}
\newcommand{\by}{\ensuremath{\mathbf{y}}}
\newcommand{\mP}{\ensuremath{\mathbf{P}}}
\newcommand{\mI}{\ensuremath{\mathbf{I}}}
\newcommand{\mQ}{\ensuremath{\mathbf{Q}}}
\newcommand{\cX}{\ensuremath{\mathcal{X}}}
\newcommand{\cY}{\ensuremath{\mathcal{Y}}}
\newcommand{\cL}{\ensuremath{\mathcal{L}}}

\iccvfinalcopy 

\ificcvfinal\fi

\begin{document}

\setlength{\abovedisplayskip}{3pt}
\setlength{\belowdisplayskip}{3pt}

\title{
Multi-Target Adversarial Frameworks for Domain Adaptation\\ in Semantic Segmentation
}

\author{Antoine Saporta$^{1,2}$ \and Tuan-Hung Vu$^2$ \and Matthieu Cord$^{1,2}$ \and Patrick Pérez$^2$ \and 
$^1$Sorbonne University \and
$^2$Valeo.ai \and
{\tt\small \{antoine.saporta, tuan-hung.vu, matthieu.cord, patrick.perez\}@valeo.com}
}

\twocolumn[{
	\renewcommand\twocolumn[1][]{#1}
	\maketitle
	\centering
 	\vspace{-0.2cm}
	\includegraphics[width=0.95\linewidth]{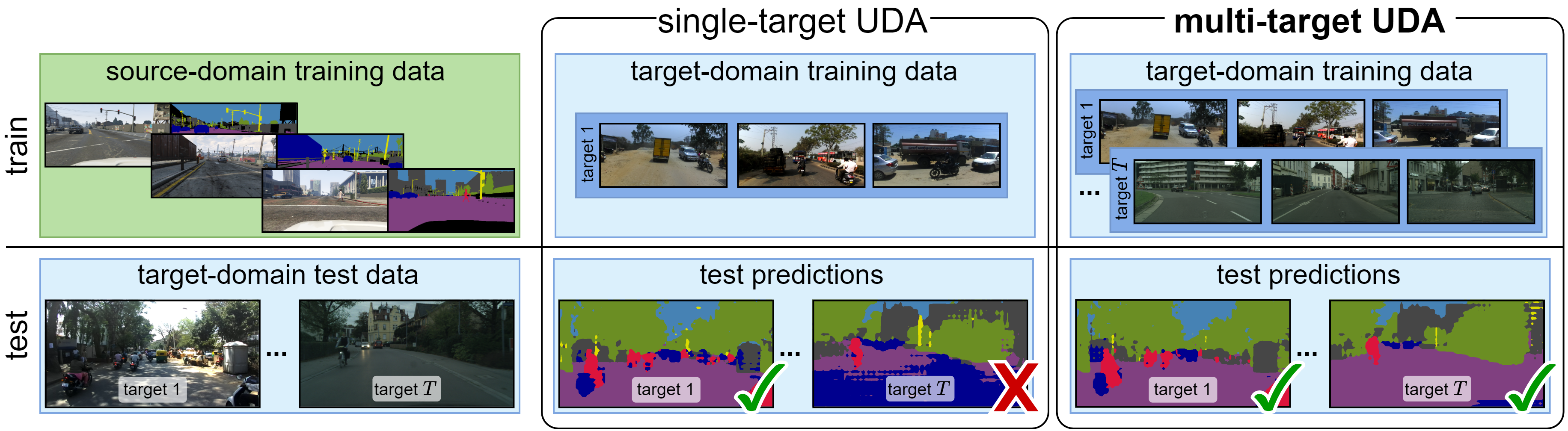}
	\vspace{-0.2cm}
	\captionof{figure}
	{\textbf{Multi-target unsupervised domain adaptation (UDA) for semantic segmentation.} In the standard single-target setting, UDA methods produce good segmentation in the target domain they are trained on, but generalize poorly to other unseen domains. Multi-target UDA aims at excelling in the multiple domains the model is trained on. 
	(\textit{Top}) The information available during training is composed of source-domain RGB images with ground-truth semantic maps (green), here from GTA5, and unannotated RGB images from target domain(s) (blue), here from IDD (`target 1') and Cityscapes (`target $T$'). (\textit{Bottom}) Test-time segmentation is on new images from the target domains, without knowing which domain they stem from.
    }
	\label{fig:teaser}
    \vspace{0.4cm}
}]

\ificcvfinal\thispagestyle{empty}\fi

\begin{abstract}
In this work, we address the task of unsupervised domain adaptation (UDA) for semantic segmentation in presence of multiple target domains: The objective is to train a single model that can handle all these domains at test time.
Such a multi-target adaptation is crucial for a variety of scenarios that real-world autonomous systems must handle. 
It is a challenging setup since one faces not only the domain gap between the labeled source set and the unlabeled target set, but also the distribution shifts existing within the latter among the different target domains.
To this end, we introduce two adversarial frameworks: (i) \emph{multi-discriminator}, which explicitly aligns each target domain to its counterparts, and (ii) \emph{multi-target knowledge transfer}, which learns a target-agnostic model thanks to a multi-teacher/single-student distillation mechanism.
The evaluation is done on four newly-proposed multi-target benchmarks for UDA in semantic segmentation.
In all tested scenarios, our approaches consistently outperform baselines, setting competitive standards for the novel task.
\end{abstract}

\section{Introduction}

Recent advances in domain adaptation help alleviate the labeling efforts required for training fully-supervised models, which is especially helpful for tasks like semantic segmentation.
Most previous works address the single-target setting whose goal is to adapt from source to a particular target domain of interest, e.g. a specific urban area.
However in practice, the perception system is often put to test in various scenarios including different cities, weathers or lighting conditions.
To deal with multiple test distributions, one can straight-forwardly adopt single-target techniques by either (i) training multiple models for all target domains and adaptively activating one at test time or (ii) merging all target data and treat them as being drawn from a single target distribution.
While the former strategy raises storage issues for embedded platforms and is difficult to scale up, the latter overlooks distribution shifts across different target domains.

In this work, we address multi-target unsupervised domain adaptation (UDA) in semantic segmentation.
We aim to learn a single segmenter that achieves equally good performance in all target domains, simultaneously closing distribution gaps between labeled-unlabeled data (source \vs target) and among target domains (target \vs target).
Our work is inline with recent efforts~\cite{chen2019blending,gholami2020unsupervised,Liu_2020_CVPR} toward more practical domain adaption settings for real-life applications.
Different from most existing multi-target works that specifically consider image classification, we study here the more complex task of semantic segmentation.

We propose two adversarial UDA frameworks with architectures and learning schemes designed for the multi-target setup.
The \emph{multi-discriminator} model explicitly reduces both source-target and target-target domain gaps via adversarial learning -- each target domain is aligned to its counterparts.
Our second framework, called \emph{multi-target knowledge transfer} (MTKT) relaxes the multi-target optimization complexity by adopting a multi-teacher/single-student mechanism.
Each \emph{target-specific} teacher handles a specific source-target domain gap via adversarial training; The \emph{target-agnostic} student is learned from all teachers to achieve target-target alignment and to perform equally well in all target domains.

Our contributions can be summarized as follows:
\begin{itemize}[itemsep=0pt, parsep=3pt, topsep=3pt, leftmargin=15pt]
   \item We propose two multi-target UDA frameworks for semantic segmentation.
    \item We define four different evaluation benchmarks for the task making use of existing semantic segmentation datasets, i.e. GTA5~\cite{richter-eecv2016}, Cityscapes~\cite{cordts-cvpr2016}, Mapillary Vistas~\cite{neuhold-iccv2017} and India Driving Dataset~\cite{varma-wacv2019}.
    \item We conduct extensive experiments of these two models against state-of-the-art baselines on the proposed benchmarks. Our approaches report consistent improvements over addressed baselines.
\end{itemize}

\section{Related Works}

\noindent{\textbf{Unsupervised Domain Adaptation for Semantic Segmentation.~}}
UDA is a setting that has received a lot of attention recently~\cite{hoffman2017cycada,long2015learning,sun2016deep,tsai-cvpr2018,vu-cvpr2019,wu2018dcan}. The objective is to train a model on an unlabeled \emph{target} domain by leveraging information from a labeled \emph{source} domain, which is usually performed by aligning in some way the distributions between source and target domains. Some strategies include constraining the training with regularization such as maximum mean discrepancy (MMD)~\cite{long2015learning} or correlation alignment~\cite{sun2016deep}. Most recent works, in particular in UDA for semantic segmentation, adopt an adversarial training strategy either at feature level~\cite{hoffman-arxiv2016} or output level~\cite{tsai-cvpr2018,vu-cvpr2019}. Some works also include a form of style transfer or image translation~\cite{hoffman2017cycada,wu2018dcan,yang2020fda} to obtain target-looking source images while keeping source annotation.
Additionally, a few works resort to ``pseudo-labeling'' ~\cite{li-cvpr2019,saporta-cvpr2020,zou2018unsupervised} to 
refine their model with the help of automatically produced annotation in the target domain.

While these methods are really effective to adapt from one domain to another, their UDA setting is limited. In real-world scenarios, data may come from various domains: In urban scenes for instance, such domain variations may stem from different sensors, weather conditions or cities. While the underlying distribution is similar across domains, traditional UDA models are not robust to changes of target domains. Moreover, since they are specifically designed for single-source to single-target alignment, they fail to leverage information across more source or target domains.

Some recent works extend the standard UDA setting in semantic segmentation to more source or target domains. MADAN~\cite{zhao2019multi} tackles the task of multi-source domain adaptation for semantic segmentation where a model is trained using multiple labeled source domains and adapted on a single target domain. The authors first transform source images into adapted domains, similar to the target domain, then bring these new domains closer together with a sub-domain aggregation discriminator. They finally train the segmentation network by performing adversarial feature-level alignment between adapted and target domains. Closer to our setting, OCDA~\cite{Liu_2020_CVPR} addresses UDA with an \emph{open compound} target domain: In this task, the target domain may be considered as a combination of multiple homogeneous target domains -- for instance, similar weather conditions such as `sunny', `foggy', etc. -- where the domain labels are not known during training. Moreover, previously unseen target domains may be encountered during inference. Unlike OCDA, our multi-target setting assumes that the domain of origin is known at training time and that no new domains are faced at test time (except in additional generalization experiments).

\smallskip\noindent{\textbf{Multi-Target Domain Adaptation for Classification.}~}
Multi-target domain adaptation is still a fairly recent setting in the literature and 
 mostly tackles classification tasks.
Two main scenarios emerge in the works on this task. In the first one, even though the target is considered composed of multiple domains with gaps and misalignments, the domain labels are unknown during training and test. 
\cite{peng2019domain} proposes an architecture that extracts domain-invariant features by performing source-target domain disentanglement. 
Moreover, it also removes class-irrelevant features by adding a class disentanglement loss. 
In a similar setting, 
\cite{chen2019blending} presents an adversarial meta-adaptation network that both aligns source with mixed-target features and uses an unsupervised meta-learner to cluster the target inputs into $k$ clusters, which are adversarially aligned. 
In the second scenario, the target identities are labeled on the training samples but remain unknown during inference. 
To handle it, \cite{yu2018multi} learns a common parameter dictionary from the different target domains and extracts the target model parameters by sparse representation; 
\cite{gholami2020unsupervised} adopts a disentanglement strategy by capturing separately both domain-specific private features and feature representations by learning a domain classifier and a class label predictor, and trains a shared decoder to reconstruct the input sample from those disentangled representations.

In the present work, we adopt the second multi-target hypothesis: The target identities are known for the training samples but not for test ones.
In fact, assuming that this information is available at test time is incompatible with some practical scenarios. More importantly, it would hinder generalization to previously-unseen domains, an important issue for autonomous systems in the wild.
To the best of our knowledge, tackling semantic segmentation in this multi-target UDA scenario has 
only been proposed in a recently published concurrent work~\cite{isobe2021mtda}.
This work proposes to train a fully-fledged segmentation network for each domain and to ensure consistency among these multiple networks with image stylization between domains.

\section{Adversarial Adaptation to Multiple Targets}

\subsection{Problem Formulation}

\noindent{\textbf{Standard Unsupervised Domain Adaptation.}~}
The standard setting that is addressed in most UDA works is single source and single target.
For adaptation, the model is trained on both a source-domain set $\cX_{\text{s}}$ with the associated ground-truth set $\cY_{\text{s}}$ and an unlabeled target-domain set $\cX_{\text{t}}$.

For semantic segmentation in $C$ classes, sets $\cX_{\text{s}}$ and $\cX_{\text{t}}$ contain training images $\bx \in\mathbb{R}^{H\times W \times 3}$, while the annotation set  $\cY_{\text{s}}\subset [0,1]^{H\times W \times C}$ contains for each $\bx\in\cX_{\text{s}}$ a map $\by$ of $H\times W$ one-hot vectors indicating the ground-truth semantic classes for all pixels.

A segmentation network $F$ takes an image $\bx$ as input and predicts a soft-segmentation map $[\mP_{\bx}(\mv k)]_{\mv k\in [H]\!\times\![W]\!\times\![C]}$.\footnote{We use notation $[A]= \left\{1,\ldots,A\right\}$ for $A\in\mathbb{N}^*$.} The final segmentation map, $F(\bx)$, is given by max-score class,  $\arg\max_{c\in [C]} \mP_{\bx}(i,j,c)$, at each pixel. 
UDA methods aim at aligning the distributions of the source-domain 
and target-domain training data
such that, at test time, the segmenter $F$ produces satisfactory predictions for target-domain inputs, without having been trained on labeled images from this domain.

\smallskip\noindent{\textbf{Multi-Target UDA.}~}
In this work, we consider a different UDA scenario where $T\geq 2$ distinct target domains must be jointly handled. These target domains are represented by unlabeled training sets $\cX_{\text{t},n} \subset \mathbb{R}^{H\times W \times 3}$, $n\in [T]$. 
Similar to the standard setting, we assume that the annotated training examples $(\bx,\by)\in \cX_{\text{s}}\times\cY_{\text{s}}$ stem from a single source domain, a specific synthetic environment for instance.
The main goal is to train a single segmenter $F$ that achieves equally good results on all target-domain test sets. 
While the target domain of origin is known for all unlabeled training examples, we assume as in classification approaches in \cite{gholami2020unsupervised,yu2018multi} that this information is not accessible at test time.

\begin{figure}[t!]
    \centering
    \includegraphics[trim=0 0 14mm 0,clip,width=.48\textwidth]{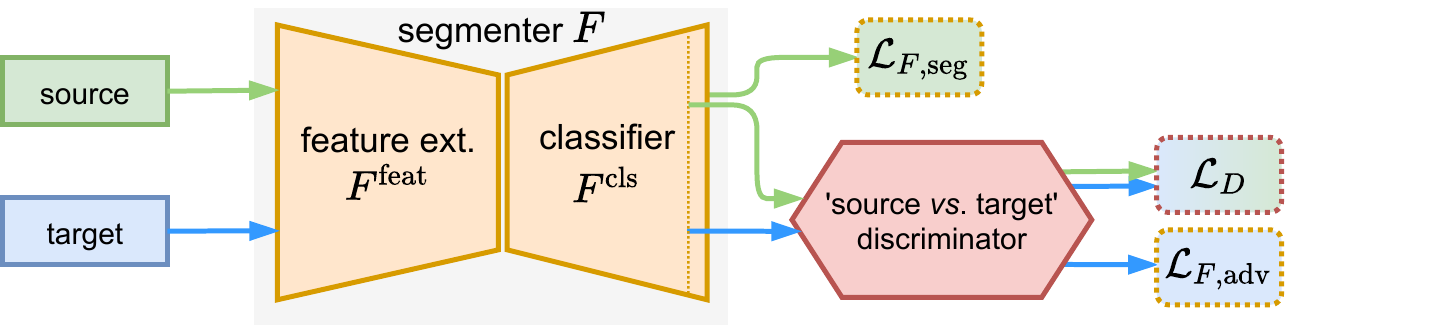}
    \caption{
     \textbf{Training in adversarial UDA.} The segmentation model under training ingests source-domain (green) and target-domain (blue) data. The former contribute to the segmentation loss, the latter to the adversarial loss, and both to the discriminator's loss. The three losses (dotted boxes) are defined in Eqs.\,\eqref{eq:l_d} and \eqref{eq:l_f}.}
    \label{fig:uda}
\end{figure}

\subsection{Revisiting Adversarial UDA Approach}

Recent state-of-the-art single-target UDA approaches are based on adversarial training to align source-target distributions.
In such approaches, besides the segmenter $F$ with parameters $\theta$, an additional network $D$ with parameters $\phi$, called discriminator, is trained to play the segmenter's ``adversary'':
$D$ is learned to predict the domain of an input from suitable representations extracted by $F$ such as intermediate or close-to-output features.  
Concurrently, $F$ tries to produce results that can fool $D$ into wrong discrimination.
In semantic segmentation, adversarial approaches operating on close-to-prediction representations have the most success. 
AdaptSegnet~\cite{tsai-cvpr2018} proposes to have adversarial learning on top of the soft-segmentation predictions $\mP_{\bx}$. AdvEnt~\cite{vu-cvpr2019} improves AdaptSegnet by using instead the ``weighted self-information'' maps $\mI_{\bx}$,\footnote{Defined as $\mI_{\bx} = -\mP_{\bx} \,\log \mP_{\bx}$, with entry-wise operations.} which brings additional entropy-minimization effect through adversarial alignment.
Such single-target adversarial frameworks serve as the building block on top of which we develop our multi-target strategies.
Hereafter, we denote $\mQ_{\bx}$ the used representation, which stands for either $\mP_{\bx}$ in~\cite{tsai-cvpr2018} or $\mI_{\bx}$ in~\cite{vu-cvpr2019}.

In practice, $D$ is a fully-convolutional binary classifier with parameters $\phi$. 
It classifies segmenter's output $\mQ_{\bx}$ into either class $1$ (source) or $0$ (target).
To train the discriminator, we minimize the classification loss: 
\begin{equation}
    \cL_D(\phi) = 
    \big\langle \cL_\text{BCE}(D(\mQ_{\bx}),1) \big\rangle_{\cX_{\text{s}}} + 
    \big\langle \cL_\text{BCE}(D(\mQ_{\bx}),0) \big\rangle_{\cX_{\text{t}}},
    \label{eq:l_d}
\end{equation}
where $\cL_\text{BCE}$ stands for the binary cross-entropy loss and $\langle . \rangle$ denotes averaging over the set in subscript.

Concurrently, the segmenter $F$ is trained over its parameters $\theta$ not only to minimize the supervised segmentation loss $\cL_{F,\text{seg}}$ on source-domain data, but also to fool the discriminator $D$ via minimizing an adversarial loss $\cL_{F,\text{adv}}$.
The final objective reads:
\begin{equation}
    \cL_{F}(\theta) = 
    \underbrace{\big\langle \cL_\text{CE}(\bx,\by) \big\rangle_{\cX_{\text{s}}}}_{\cL_{F,\text{seg}}(\theta)} + \lambda_\text{adv}
    \underbrace{\big\langle \cL_\text{BCE}(D(\mQ_{\bx}),1) \big\rangle_{\cX_{\text{t}}}}_{\cL_{F,\text{adv}}(\theta)},
    \label{eq:l_f}
\end{equation}
with a weight $\lambda_\text{adv}$ balancing the two terms; $\cL_\text{CE}$ is the common  cross-entropy loss.
During training, one alternately minimizes the two losses $\cL_{D}$ and $\cL_{F}$.

Figure\,\ref{fig:uda} provides a high-level view of the training flow in recent adversarial UDA approaches. 
For more details, we refer the readers to \cite{tsai-cvpr2018,vu-cvpr2019} for instance.
To later facilitate the presentation of our proposed strategies, the segmenter $F$ is decoupled into a feature extractor, $F^{\mathrm{feat}}$, followed by a pixel-wise classifier, $F^{\mathrm{cls}}$.

\smallskip\noindent{\textbf{Discussion.}~}
Approaches like \cite{tsai-cvpr2018,vu-cvpr2019} 
handle only one source domain and one target domain. 
In our setting with multiple target domains, a simple strategy is to merge all target datasets into a single one and then to utilize an existing single-source single-target UDA framework.
Such a strategy however disregards the inherent discrepancy among target domains. 
As we show in the experiments, this multi-target baseline is less effective than the proposed strategies which explicitly handle inter-target domain shifts. 
In what follows, we describe these two novel frameworks.

\subsection{Multi-Target Frameworks}
\begin{figure}[t!]
    \centering
    \includegraphics[trim=0 0 18mm 0,clip,width=.48\textwidth]{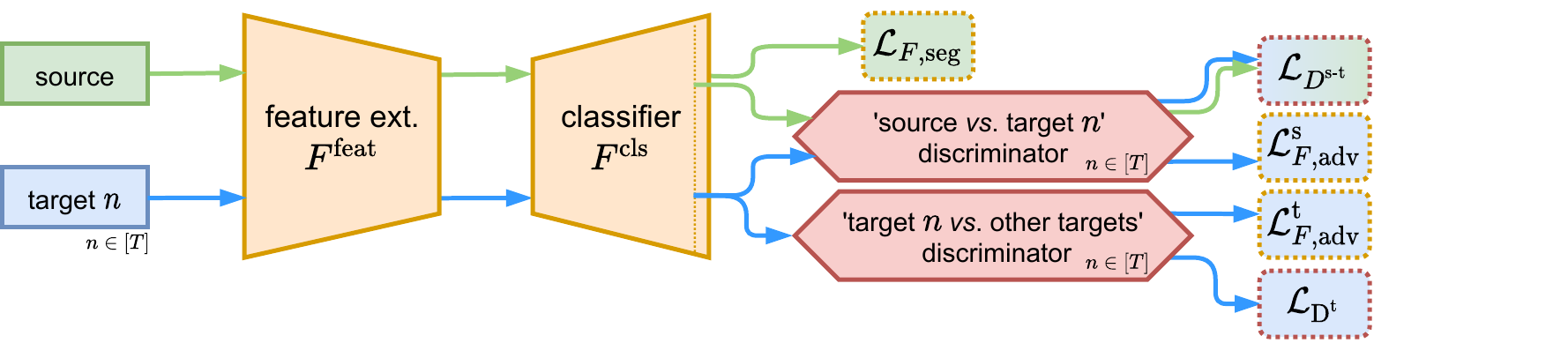}
    \caption{\textbf{Multi-discriminator approach to multi-target UDA}. With Multi-Dis., the segmenter is trained against two types of adversaries that discriminate respectively source \textit{vs}. one target and one target \textit{vs}. all other targets. The four types of adversarial losses are defined in Eqs.\,\eqref{eq:l_ds}, \eqref{eq:l_fsadv}, \eqref{eq:l_dt} and \eqref{eq:l_ftadv}.
    Symbols and colors follow those in Figure~\ref{fig:uda}.}
    \label{fig:mdisc}
\end{figure}

\noindent{\textbf{Multi-Discriminator.}~}
Our first strategy for multi-target UDA, called \textit{multi-discriminator} (`Multi-Dis.' in short), relies on two types of discriminators to align each target domain with the source (source-target discriminators) and with other targets (target-target discriminators). Figure~\ref{fig:mdisc} illustrates this first approach.

\noindent\textit{Source-target adversarial alignment}.~ 
We introduce a discriminator $D^{\text{s-t}}_{n}$ with parameters $\phi^{\text{s-t}}_{n}$ for each target domain $n$. 
It is learned to discriminate $\cX_{\text{t},n}$ from the source set $\cX_{\text{s}}$.
By denoting $\cL_{D^{\text{s-t}}_{n}}$ the minimization objective of this discriminator, defined as in \eqref{eq:l_d} on domain $n$, we train these $T$ source-target discriminators with the mean objective:
\begin{equation}
    \cL_{D^\text{s-t}}\big(\phi^{\text{s-t}}_{1:T}\big) = \frac{1}{T}\sum\limits_{n\in[T]} \cL_{D^{\text{s-t}}_{n}}\big(\phi^{\text{s-t}}_{n}\big).
    \label{eq:l_ds}
\end{equation}

Concurrently, the segmenter $F$ is trained to fool these $T$ discriminators with the adversarial objective: 
\begin{equation}
    \cL^{\text{s}}_{F,\text{adv}}(\theta) = \frac{1}{T}\sum\limits_{n\in[T]}
    \big\langle \cL_{\text{BCE}}(D^{\text{s-t}}_n(\mQ_{\bx}),1) \big\rangle_{\cX_{\text{t},n}}.
    \label{eq:l_fsadv}
\end{equation}

\noindent\textit{Target-target adversarial alignment}. 
In the above source-target alignment, the source acts as an anchor for each target to ``pull'' closer the other targets. However, as this alignment is imperfect, there remain gaps across targets, which we propose to reduce further by additional target-target alignments. 
To this end, we introduce for each target domain $n$ a discriminator $D^{\text{t}}_{n}$ with parameters $\phi^{\text{t}}_{n}$ that classifies $\cX_{\text{t},n}$ (class 1) \vs{} all other target domains $\cX_{\text{t},k},\,k\neq n$ (class 0), resulting in $T$ 1-\vs-all discriminators.
The target-target discriminator $D^{\text{t}}_n$ is trained by minimizing the loss
\begin{equation}
    \resizebox{\columnwidth}{!}{$
    \cL_{D^{\text{t}}_{n}}\big(\phi^{\text{t}}_{n}\big) = 
    \big\langle \cL_{\text{BCE}}(D^{\text{t}}_{n}(\mQ_\bx),1) \big\rangle_{\cX_{\text{t},n}} \!\!+ 
    \big\langle \cL_{\text{BCE}}(D^{\text{t}}_{n}(\mQ_\bx),0) \big\rangle_{\!\mathop{\bigcup}\limits_{k\neq n}\!\!\cX_{\text{t},k}}.
    $}
\end{equation}

\begin{figure}[t!]
    \centering
     \includegraphics[trim=0 0 12mm 0,clip,width=.48\textwidth]{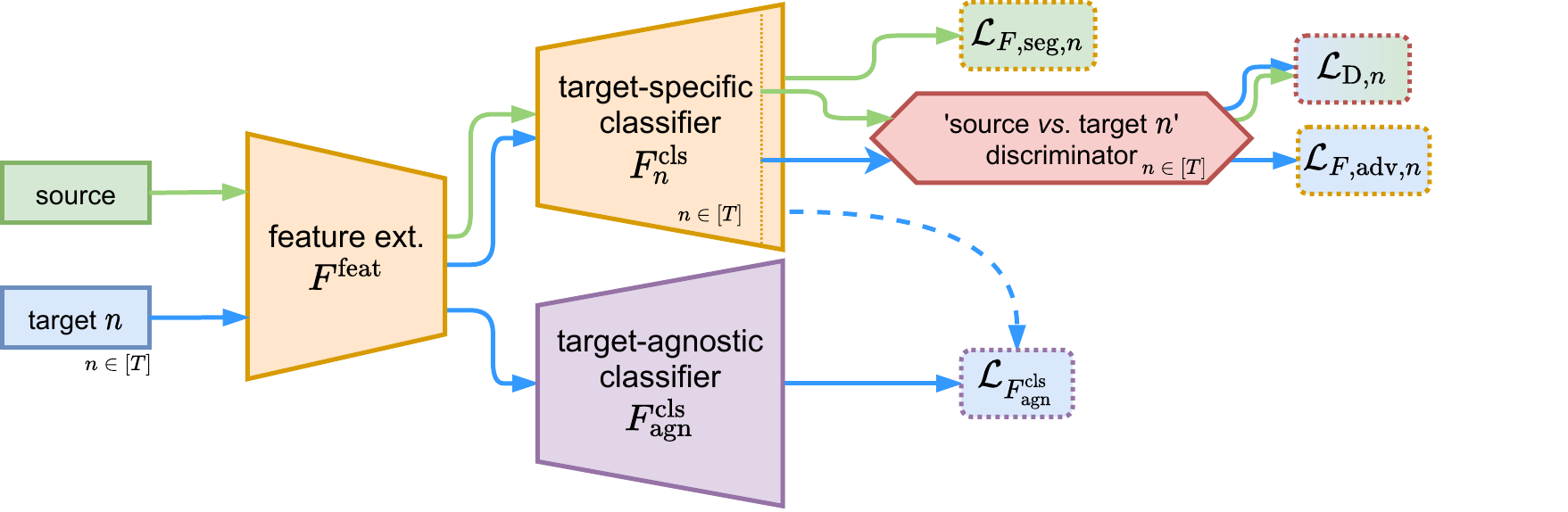}
    \caption{\textbf{Multi-target knowledge transfer approach to multi-target UDA.} 
    With MTKT, a set of target-specific segmenters is first trained adversarially. Their knowledge is then jointly distilled to the target-agnostic segmenter whose loss \eqref{eq:l_fagn} is not back-propagated into the target-specific branches (as indicated by the dotted arrow). Symbols and colors follow those in Figure~\ref{fig:uda}.}
    \label{fig:mclf}
\end{figure}

The collective objective of all target-target discriminators now reads: 
\begin{equation}
    \cL_{D^{\text{t}}}\big(\phi^{\text{t}}_{1:T}\big) = \frac{1}{T}\sum\limits_{n\in[T]} \cL_{D^{\text{t}}_{n}}\big(\phi^{\text{t}}_{n}\big).
    \label{eq:l_dt}
\end{equation}

The segmenter $F$ tries to fool all the target-target discriminators by minimizing the adversarial loss: 
\begin{equation}
    \cL^{\text{t}}_{F,\text{adv}}(\theta) = \frac{1}{T}
    \sum\limits_{n\in[T]}
    \big\langle \cL_{\text{BCE}}(D^{\text{t}}_{n}(\mQ_{\bx}),1)  \big\rangle_{\!\mathop{\bigcup}\limits_{k\neq n}\!\!\cX_{\text{t},k}}.
    \label{eq:l_ftadv}
\end{equation}

To sum up, the segmenter $F$ is trained by minimizing over $\theta$ the objective:
    \begin{equation}
        \cL_F = \cL_{F,\text{seg}} + \lambda_\text{adv}^{\text{s}} \cL^{\text{s}}_{F,\text{adv}} + \lambda_\text{adv}^{\text{t}} \cL^{\text{t}}_{F,\text{adv}},
        \label{eq:l_fmdm}
    \end{equation}
with weights $\lambda_\text{adv}^{\text{s}}$ and $\lambda_\text{adv}^{\text{t}}$ to balance the adversarial terms.

\smallskip\noindent{\textbf{Multi-Target Knowledge Transfer.}~}
The main driving force in prediction-level adversarial approaches~\cite{tsai-cvpr2018,vu-cvpr2019} 
is the adjustment of the decision boundaries. 
Alignment in feature space then follows to comply with adjusted boundaries.
We thus stress the importance of classifier design in the multi-target UDA scenario.
In our multi-discriminator approach, one classifier simultaneously handles multiple domain shifts, either source-target or target-target.
The main challenge is the instability of adversarial training, which is amplified if several adversarial losses are jointly minimized. 
Such an issue is particularly problematic in the early training phase when most target predictions are very noisy.
To address this challenge, we propose the \textit{multi-target knowledge transfer} (MTKT) framework, with novel network design and learning scheme which do not rely on the joint minimization of multiple adversarial losses over the same classifier module, hopefully reducing the instability of the training. 
Figure~\ref{fig:mclf} shows the MTKT architecture.

The classification part of the network is first re-designed with $T$ \emph{target-specific} instrumental classifiers, $F^{\mathrm{cls}}_n,~n\in [T]$, based on the same feature extractor $F^{\mathrm{feat}}$, each handling one specific source-target domain shift. 
Such an architecture allows separate output-space adversarial alignment for each specific source-target pair, alleviating the instability problem.
For each target-specific classifier $F^{\mathrm{cls}}_n$, we introduce a domain discriminator ${D^{\text{t}}_n}$ as to classify source \vs target $n$.
The training objectives are similar to those used in single-target models (Eqs.\,\ref{eq:l_d} and \ref{eq:l_f}). 

We then introduce a \emph{target-agnostic} classification branch $F^{\mathrm{cls}}_{\text{agn}}$ that fuses all the knowledge transferred from the $T$ target-specific classifiers. This target-agnostic classifier is the final product of the approach, \ie, the one used at test time when domain knowledge is not available.

The knowledge from the $T$ ``teachers'' is transferred to the target-agnostic ``student'' via minimizing the Kullback-Leibler divergence~\cite{hinton2015distilling} between teachers' and student's predictions on target domains.
In details, for a given sample $\bx\in{\cX_{\text{t},n}}$, we compute the KL loss
\begin{equation}\cL_{\text{KL},n}(\bx) = \sum_{\mv k\in [H]\!\times\![W]\!\times\![C]}\mP_{n,\bx}(\mv k) \log \frac{\mP_{n,\bx}(\mv k)}{\mP_{\bx}(\mv k)},
\end{equation}
where $\mP_{n,\bx}$ and $\mP_{\bx}$ are soft-segmentation predictions coming from the target-specific $F^{\mathrm{cls}}_n$ and the target-agnostic $F^{\mathrm{cls}}_{\text{agn}}$ respectively.
The minimization objective of the target-agnostic classifier $F^{\mathrm{cls}}_{\text{agn}}$ over the segmenter's parameters (including feature extractor's) then reads:
\begin{equation}
    \cL_{F^{\mathrm{cls}}_{\text{agn}}}(\theta) = 
    \frac{1}{T}\sum_{n\in[T]}
    \big\langle
    \cL_{\text{KL},n}(\bx) \big\rangle_{\cX_{\text{t},n}}.
    \label{eq:l_fagn}
\end{equation}

Minimizing KL losses 
helps $F^{\mathrm{cls}}_{\text{agn}}$ adjust its decision boundaries toward good behavior in all $T$ target domains. 
As the KL loss is back-propagated through the feature extractor,  
such an adjustment results in implicit alignment in target feature space, which overall mitigates the distribution shifts between the $T$ domains.

\smallskip\noindent{\textbf{Discussion.}~} Unlike Multi-Dis., the multi-teacher/single-student mechanism in MTKT avoids direct alignment between unlabeled parts.  
The target-agnostic classifier is encouraged to adjust its decision boundaries to favor all the target-specific teachers, thus helping cross-target alignment.

Although we build our frameworks over output-space alignment~\cite{vu-cvpr2019,tsai-cvpr2018}, note that they could be adapted to other adversarial feature-alignment methods~\cite{hoffman-arxiv2016}. Moreover, orthogonal approaches like pseudo-labeling can also be included in our frameworks and we show some experiments with such an addition in Section~\ref{sec:further_exp}.

\section{Experiments}
\label{sec:exp}

\subsection{Experimental Details}

\smallskip\noindent{\textbf{Datasets.}~}
We build our experiments on four urban driving datasets, one being synthetic and the three others being recorded in various geographic locations:
\begin{itemize}[itemsep=0pt]
    \item GTA5~\cite{richter-eecv2016} is a dataset of 24,966 labeled synthetic images 
    generated from the eponymous video game;
    \item Cityscapes~\cite{cordts-cvpr2016} contains labeled urban scenes from cities around Germany, split in training and validation sets of 2,975 and 500 samples respectively;
    \item IDD~\cite{varma-wacv2019} is an Indian urban dataset having 6,993 training and 981 validation labeled scenes;
    \item Mapillary Vistas~\cite{neuhold-iccv2017} is a dataset collected in multiple cities around the world, which is composed of 18,000 training and 2,000 validation labeled scenes.
\end{itemize}
Though all containing urban scenes, the four datasets have different labeling policies and semantic granularity.
We follow the protocol used in~\cite{lee-iclr2019,vu-iccv19} and standardize the label set with 7 super classes, common to all four datasets: \emph{flat}, \emph{construction}, \emph{object}, \emph{nature}, \emph{sky}, \emph{human} and \emph{vehicle}. The mapping from original classes to these super classes is given in the Supplementary Material.

When Cityscapes, IDD or Mapillary are used as target domain, only unlabeled images from them are used for training, by definition of the UDA problem.

\smallskip\noindent{\textbf{Implementation Details.}~}
Our experiments are conducted with PyTorch~\cite{paszke-nips2017}.
The adversarial framework is based on AdvEnt's published code.\footnote{\url{https://github.com/valeoai/ADVENT}} 
We adopt DeepLab-V2~\cite{chen-tpami2018} as the semantic segmentation model, built upon the ResNet-101~\cite{he-cvpr2016} backbone initialized with ImageNet~\cite{deng-cvpr2009} pre-trained weights.
The segmenters are trained by Stochastic Gradient Descent~\cite{bottou2010large} with learning rate $2.5\times 10^{-4}$, momentum $0.9$ and weight decay $10^{-4}$.
We train the discriminators using an Adam optimizer~\cite{kingma-iclr2015} with learning rate $10^{-4}$.  All experiments were conducted at the $640 \times 320$ resolution.

For MTKT, we ``warm up'' the target-specific branches for 20,000 iterations before training the target-agnostic branch.
The warm-up step avoids distillation of noisy target predictions in the early phase, which helps stabilize target-agnostic training.

\subsection{Main results}

We consider four setups, varying the type of domain-shift (`syn-2-real' or `city-2-city') or the number $T$ of targets (two to three domains).
To measure per-target segmentation performance, we use the standard mean Intersection-over-Union (mIoU) metric.
For multi-target performance, we report the mIoU averaged over the target domains; Using the average helps mitigate the potential bias caused by target evaluation sets with substantially different sizes.

\begin{table}[t!]
\setlength\heavyrulewidth{0.25ex}
\aboverulesep=0ex
\belowrulesep=0ex
\resizebox{.48\textwidth}{!}{%
        \begin{tabular}{l|l|c| c c c c c c c|l|l<{\kern-\tabcolsep}}
            \toprule
            \multicolumn{12}{c}{\textbf{GTA5\,$\shortrightarrow$\,Cityscapes\,$+$\,Mapillary}}\\
            \midrule
            Method & Target & Train & \rotatebox{90}{flat} & \rotatebox{90}{constr.} & \rotatebox{90}{object} & \rotatebox{90}{nature} & \rotatebox{90}{sky} & \rotatebox{90}{human} & \rotatebox{90}{vehicle\,} & mIoU &  \shortstack[l]{mIoU\\Avg.} \\
            \midrule
            \multirow{4}{*}{\shortstack[l]{Single-Target\\Baselines~\cite{vu-cvpr2019}}} & Cityscapes  & \checkmark & 93.5 & 80.5 & 26.0 & 78.5 & 78.5 & 55.1 & 76.4 & 69.8 (*)&\multirow{2}{*}{66.6}\\
            & Mapillary & - & 86.8 & 69.0 & 30.2 & 71.2 & 91.5 & 35.3 & 59.5 & 63.4$_{\down{\,6.2}}$& \\
            \cmidrule(lr){2-12}
            & Cityscapes & - & 89.3 & 79.3 & 19.5 & 76.9 & 84.6 & 47.7 & 63.0 & 65.8$_{\down{\,4.0}}$& \multirow{2}{*}{67.7} \\
            & Mapillary & \checkmark & 89.5 & 72.6 & 31.0 & 75.3 & 94.1 & 50.7 & 73.8 & 69.6 (*) &\\
            \midrule
            \multirow{2}{*}{\shortstack[l]{Multi-Target\\Baseline~\cite{vu-cvpr2019}}}& Cityscapes & \checkmark & 93.1 & 80.5 & 24.0 & 77.9 & 81.0 & 52.5 & 75.0 & 69.1$_{\down{\,0.7}}$&\multirow{2}{*}{68.9} \\
            & Mapillary & \checkmark &90.0 & 71.3 & 31.1 & 73.0 & 92.6 & 46.6 & 76.6 & 68.7$_{\down{\,0.9}}$&\\
            \midrule
            \rowcolor[gray]{.92} & Cityscapes & \checkmark & 94.5 & 80.8 & 22.2 & 79.2 & 82.1 & 47.0 & 79.0 & 69.3$_{\down{\,0.5}}$ & \\
            \rowcolor[gray]{.92} \multirow{-2}{*}{Multi-Dis.} & Mapillary & \checkmark & 89.4 & 71.2 & 29.5 & 76.2 & 93.6 & 50.4 & 78.3 & 69.8$_{\up{\,0.2}}$ & \multirow{-2}{*}{69.5} \\
            \midrule
            \rowcolor[gray]{.92} & Cityscapes & \checkmark & 95.0 & 81.6 & 23.6 & 80.1 & 83.6 & 53.7 & 79.8 & \textbf{71.1}$_{\up{\,\mathbf{1.3}}}$& \\
            \rowcolor[gray]{.92}\multirow{-2}{*}{MTKT} & Mapillary & \checkmark & 90.6 & 73.3 & 31.0 & 75.3 & 94.5 & 52.2 & 79.8 & \textbf{70.8}$_{\up{\,\mathbf{1.2}}}$& \multirow{-2}{*}{\textbf{70.9}}\\
            \bottomrule
        \end{tabular}
    }
\caption{
\textbf{Semantic segmentation performance on GTA5\,$\shortrightarrow$\,Cityscapes\,$+$\,Mapillary.} Per-class IoU (\%), per-domain mean IoU (`mIoU') and mIoU averaged over domains (`mIoU Avg.'); mIoU gain (green) or loss (red) w.r.t. corresponding per-target baselines (marked as `*'); `train': indication of the unlabeled target data used for training.
}
\label{tab:G-CM}
\vspace{-0.3cm}
\end{table} 

\smallskip\noindent{\textbf{GTA5\,$\shortrightarrow$\,Cityscapes\,$+$\,Mapillary.}~}
Table~\ref{tab:G-CM} reports segmentation results on the two target validation sets of Cityscapes and Mapillary; GTA5 is the source domain in this setup.
For comparison, we consider the single-target AdvEnt models, i.e. trained on either Cityscapes or Mapillary unlabeled images.
We have also the multi-target AdvEnt model, denoted as `Multi-Target Baseline' in Table~\ref{tab:G-CM}, which is trained on the merging of the two targets.
For all models, including the single-target ones, we report both per-target and average mIoUs.
The two rows marked with `(*)' indicate results of the single-target models on the same domains used for training, regarded as per-target baselines.

Single-target baselines achieve worse average mIoU than those trained on both domains, which indicates the benefit of having access to diverse data from multiple domains during training.
Our proposed approaches outperform the multi-target baseline with mIoU gains of $+0.6\%$ for multi-discriminator and $+2.0\%$ for MTKT.
Looking closer at the per-target results, we observe unfavorable performance if one directly transfers single-target models to a new domain.
Indeed, testing the Cityscapes-only model on Mapillary results in a drop of $-6.2\%$ mIoU compared to the reference performance and a similar drastic drop is seen for Mapillary-only model on Cityscapes.
Especially we notice important degradation on safety-critical classes like \emph{human} or \emph{vehicle} using those single-target models.
The multi-discriminator model achieves comparable mIoUs as the per-target baselines.
The MTKT model improves over the per-target baselines by significant margin, i.e. $+1.3\%$ on Cityscapes and $+1.2\%$ on Mapillary. 
Such results highlight the merit of the proposed strategies, especially MTKT.
Note that adding adversarial training on the target-agnostic branch of MTKT hinders the alignment effect, reducing the performance by $0.9\%$ mIoU\,Avg.

\begin{table}[t!]
	\setlength\heavyrulewidth{0.25ex}
	\aboverulesep=0ex
	\belowrulesep=0ex
	\resizebox{.48\textwidth}{!}{%
        \begin{tabular}{l| l | c | c c c c c c c|l | l<{\kern-\tabcolsep}}
            \toprule
            \multicolumn{12}{c}{\textbf{GTA5\,$\shortrightarrow$\,Cityscapes\,$+$\,IDD}}\\
            \midrule
            Method & Target & Train & \rotatebox{90}{flat} & \rotatebox{90}{constr.} & \rotatebox{90}{object} & \rotatebox{90}{nature} & \rotatebox{90}{sky} & \rotatebox{90}{human} & \rotatebox{90}{vehicle\,} & mIoU & \shortstack[l]{mIoU\\Avg.} \\
            \midrule
            \multirow{4}{*}{\shortstack[l]{Single-Target\\Baselines~\cite{vu-cvpr2019}}} & Cityscapes & \checkmark & 93.5 & 80.5 & 26.0 & 78.5 & 78.5 & 55.1 & 76.4 & 69.8 (*) & \multirow{2}{*}{66.5}\\
            & IDD & - & 91.3 & 52.3 & 13.3 & 76.1 & 88.7 & 46.7 & 74.8 & 63.3$_{\down{\,1.8}}$ & \\
            \cmidrule(lr){2-12}
             & Cityscapes & - & 78.6 & 79.2 & 24.8 & 77.6 & 83.6 & 48.7 & 44.8 & 62.5$_{\down{\,7.3}}$ & \multirow{2}{*}{63.8} \\
            & IDD & \checkmark & 91.2 & 53.1 & 16.0 & 78.2 & 90.7 & 47.9 & 78.9 & 65.1 (*) &   \\
            \midrule
            \multirow{2}{*}{\shortstack[l]{Multi-Target\\Baseline~\cite{vu-cvpr2019}}}& Cityscapes & \checkmark & 93.9 & 80.2 & 26.2 & 79.0 & 80.5 & 52.5 & 78.0 & 70.0$_{\up{\,0.2}}$ & \multirow{2}{*}{67.4} \\
            & IDD & \checkmark & 91.8 & 54.5 & 14.4 & 76.8 & 90.3 & 47.5 & 78.3 & 64.8$_{\down{\,0.3}}$ &  \\
            \midrule
            \rowcolor[gray]{.92} & Cityscapes & \checkmark & 94.3 & 80.7 & 20.9 & 79.3 & 82.6 & 48.5 & 76.2 & 68.9$_{\down{\,0.9}}$ &  \\
            \rowcolor[gray]{.92} \multirow{-2}{*}{Multi-Dis.}& IDD & \checkmark & 92.3 & 55.0 & 12.2 & 77.7 & 92.4 & 51.0 & 80.2 & 65.7$_{\up{\,0.6}}$ & \multirow{-2}{*}{67.3} \\
            \midrule
            \rowcolor[gray]{.92} & Cityscapes & \checkmark & 94.5 & 82.0 & 23.7 & 80.1 & 84.0 & 51.0 & 77.6 & \textbf{70.4}$_{\up{\,\mathbf{0.5}}}$ &  \\
            \rowcolor[gray]{.92} \multirow{-2}{*}{MTKT}& IDD & \checkmark & 91.4 & 56.6 & 13.2 & 77.3 & 91.4 & 51.4 & 79.9 & \textbf{65.9}$_{\up{\,\mathbf{0.8}}}$ & \multirow{-2}{*}{\textbf{68.2}} \\
            \bottomrule
        \end{tabular}%
    }
    \caption{
    \textbf{Semantic segmentation performance on GTA5\,$\shortrightarrow$\,Cityscapes\,$+$\,IDD.}
    Organization as in Tab.\,\ref{tab:G-CM}.
    }
    \label{tab:G-CI}
    \vspace{-0.2cm}
\end{table}

\smallskip\noindent{\textbf{GTA5\,$\shortrightarrow$\,Cityscapes $+$IDD.}~}
We experiment with another syn-2-real setup in which the two target datasets have noticeably different landscapes, i.e. European cities in Cityscapes and Indian ones in IDD.
Results are reported in Table~\ref{tab:G-CI}.
Here also, multi-target models outperform the single-target ones.
In this setup, the performance of Multi-Dis. is comparable to the multi-target baseline's.
We conjecture that the complex and unstable optimization problem in the multi-discriminator framework makes it difficult to achieve good alignment across targets, especially when the two targets are more noticeably different.
With a dedicated architecture and learning scheme that alleviate such an optimization issue, the MTKT model achieves the best results, in terms of both per-target and average mIoUs. 

\begin{figure*}[h]
    \centering
   {
    \begin{tabular}{p{0.1cm} c@{\hskip5pt} c@{\hskip5pt} c@{\hskip5pt} c@{\hskip5pt} c@{\hskip5pt} c@{\hskip5pt} c}
         & (a) {\footnotesize Input} & (b) {\footnotesize Ground truth} & (c) {\footnotesize City. Baseline} & (d) {\footnotesize IDD Baseline} & (e) {\footnotesize MT Baseline} & (f) {\footnotesize Multi-Dis.} & (g) {\footnotesize MTKT} \\
         \multirow{5}{*}{ \rotatebox[origin=c]{90}{\parbox[c]{4cm}{\centering {\footnotesize Cityscapes}}}}& \includegraphics[width=2.2cm]{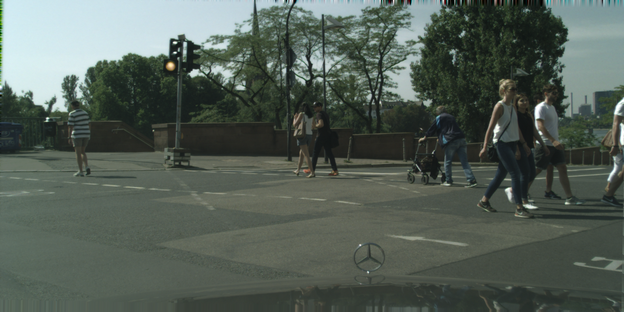} & \includegraphics[width=2.2cm]{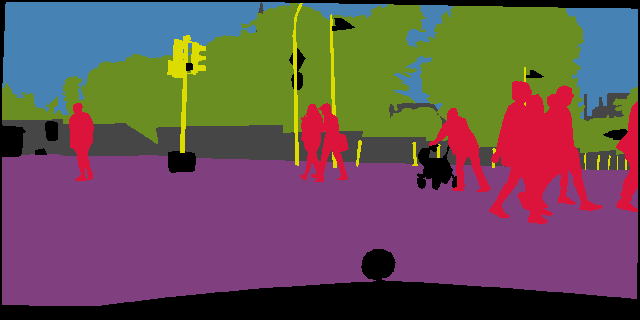} &
         \includegraphics[width=2.2cm]{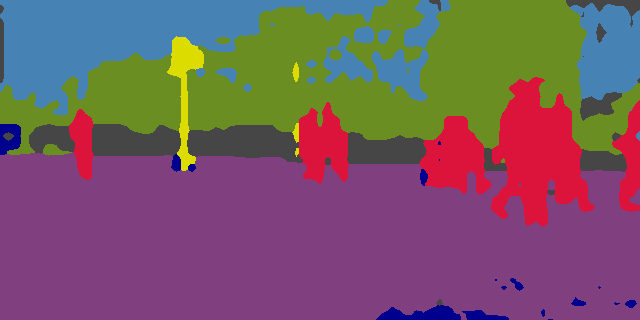} &\includegraphics[width=2.2cm]{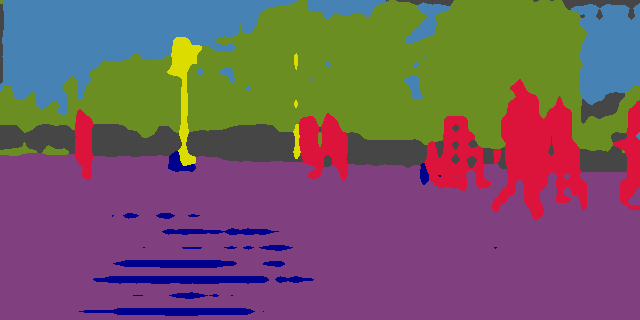} &\includegraphics[width=2.2cm]{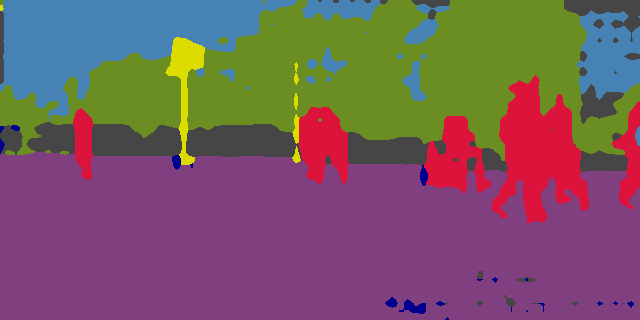} &
         \includegraphics[width=2.2cm]{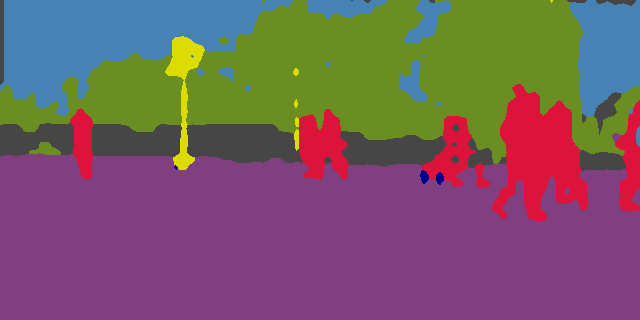} &\includegraphics[width=2.2cm]{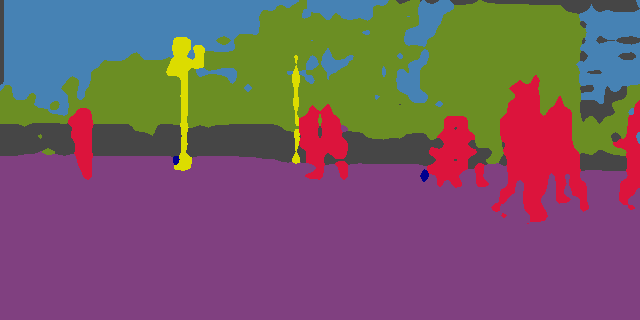}\\
      & \includegraphics[width=2.2cm]{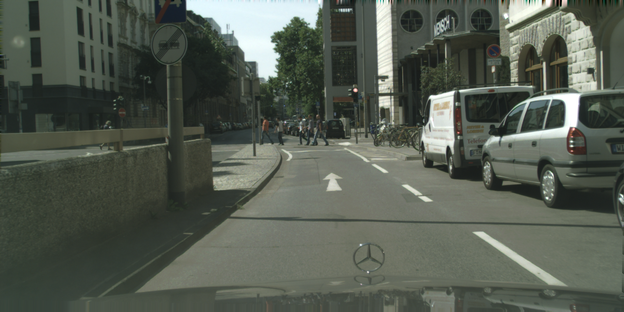} & \includegraphics[width=2.2cm]{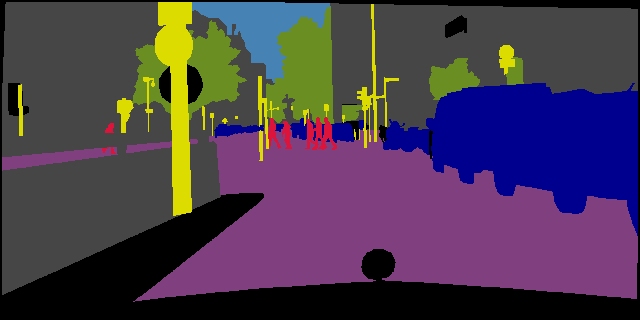} &
      \includegraphics[width=2.2cm]{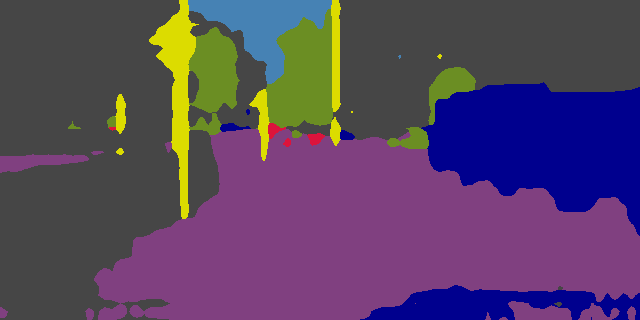}
      &
      \includegraphics[width=2.2cm]{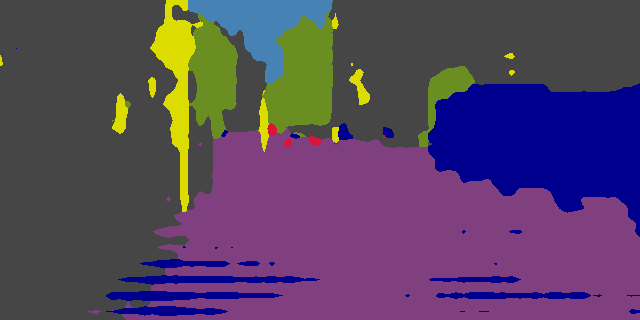}
      &\includegraphics[width=2.2cm]{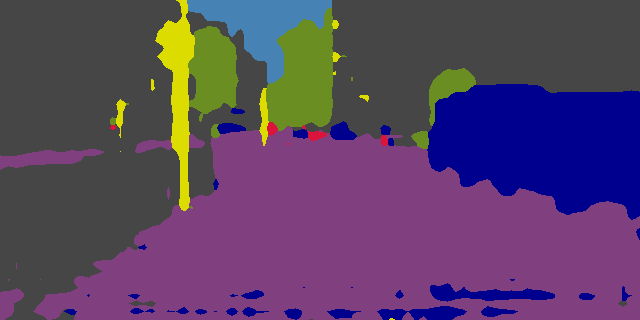} &\includegraphics[width=2.2cm]{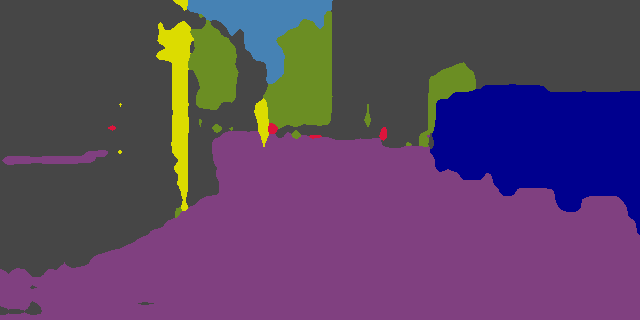} & \includegraphics[width=2.2cm]{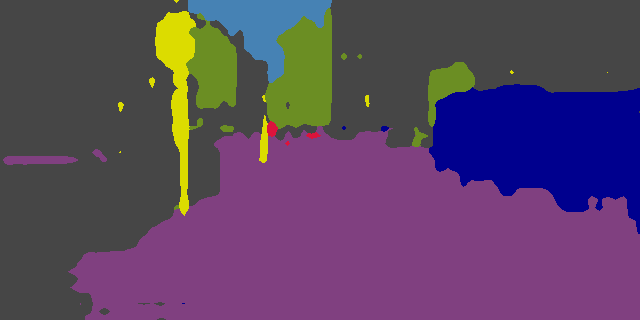}\\
        & \includegraphics[width=2.2cm]{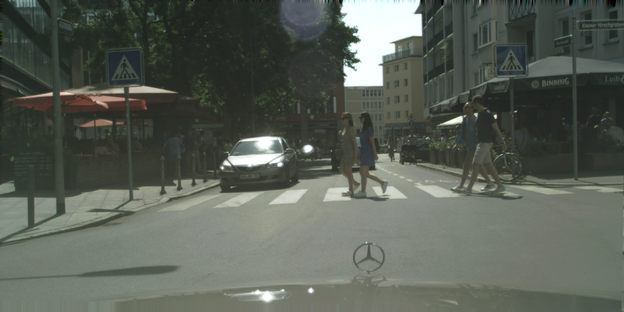} & \includegraphics[width=2.2cm]{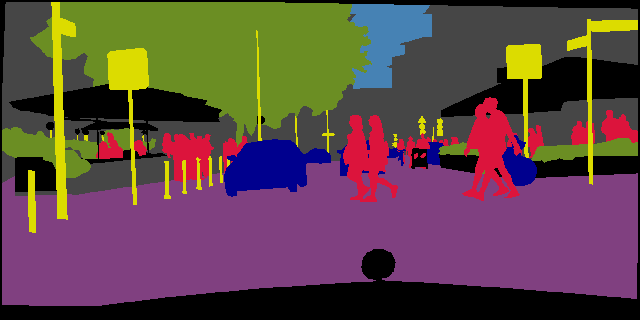} &
        \includegraphics[width=2.2cm]{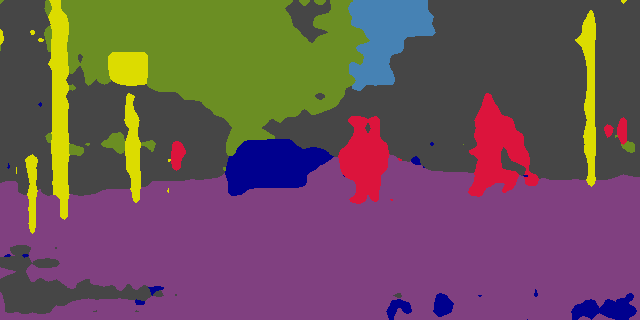} &
        \includegraphics[width=2.2cm]{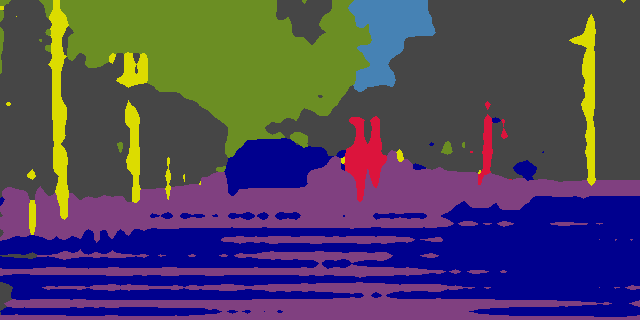}
        &\includegraphics[width=2.2cm]{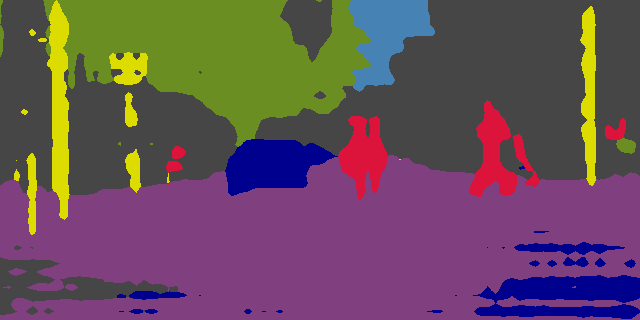}
        &\includegraphics[width=2.2cm]{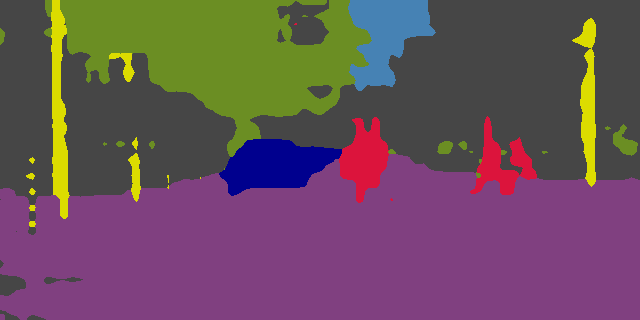} & \includegraphics[width=2.2cm]{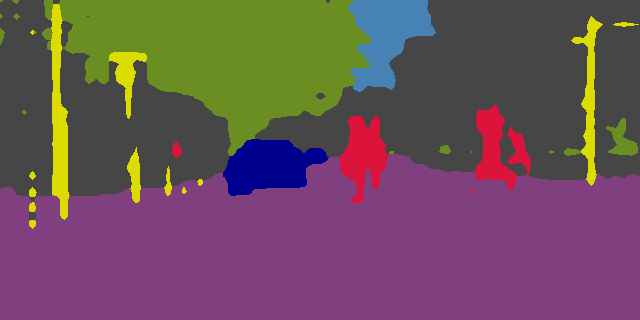}\\
        & \includegraphics[width=2.2cm]{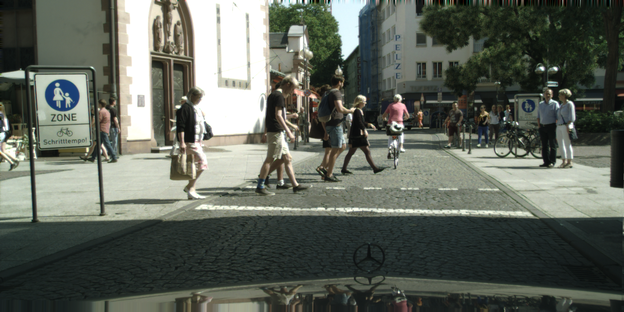} & \includegraphics[width=2.2cm]{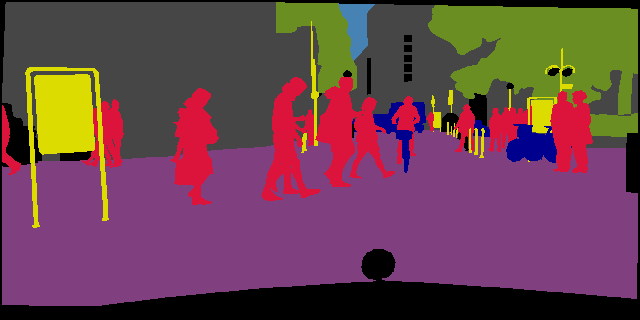} &
        \includegraphics[width=2.2cm]{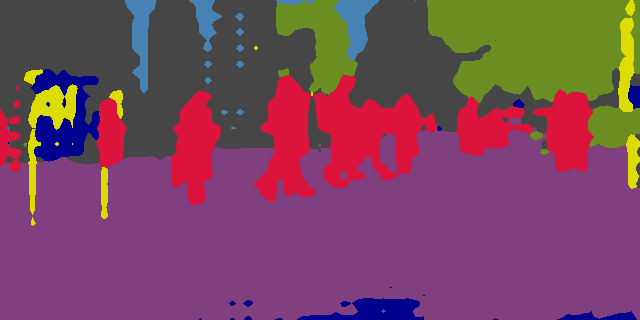} &
        \includegraphics[width=2.2cm]{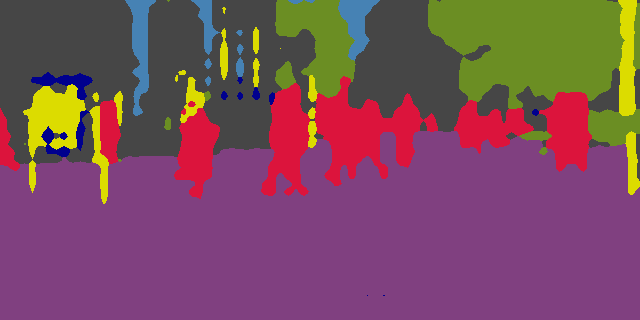} &\includegraphics[width=2.2cm]{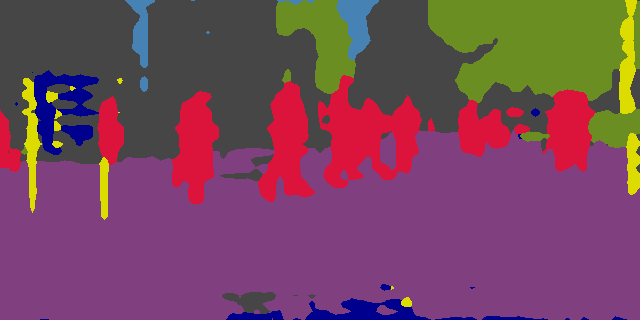} &\includegraphics[width=2.2cm]{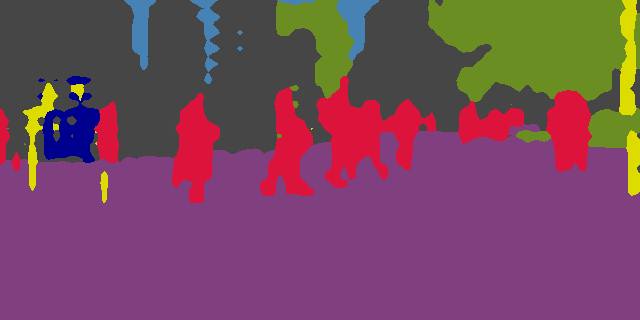} & \includegraphics[width=2.2cm]{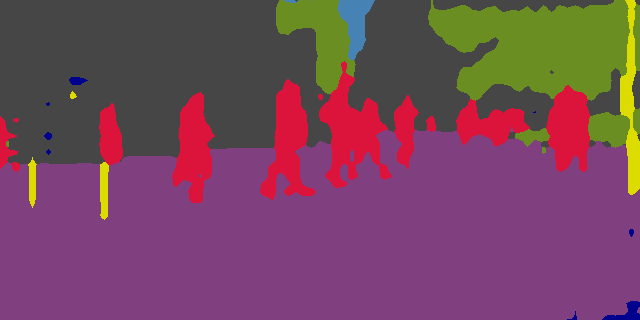}\\
        & \includegraphics[width=2.2cm]{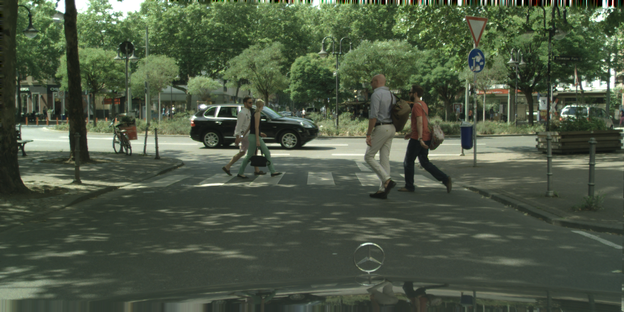} & \includegraphics[width=2.2cm]{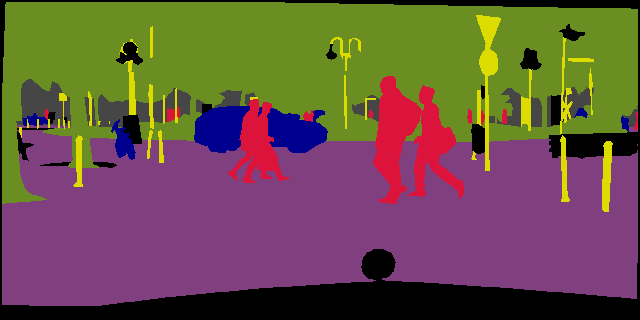} &    \includegraphics[width=2.2cm]{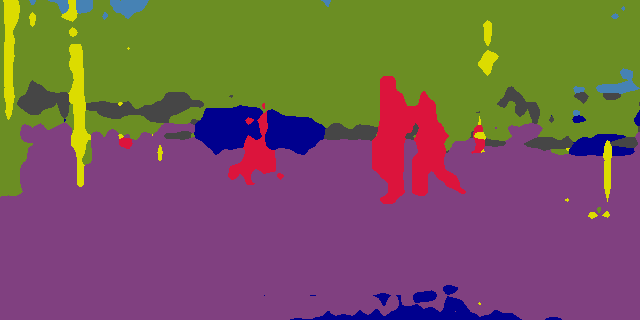} &   \includegraphics[width=2.2cm]{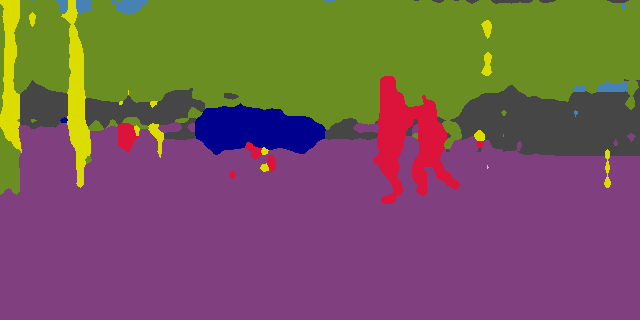} &\includegraphics[width=2.2cm]{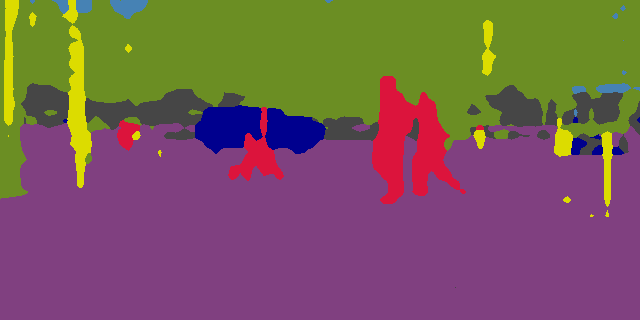} &\includegraphics[width=2.2cm]{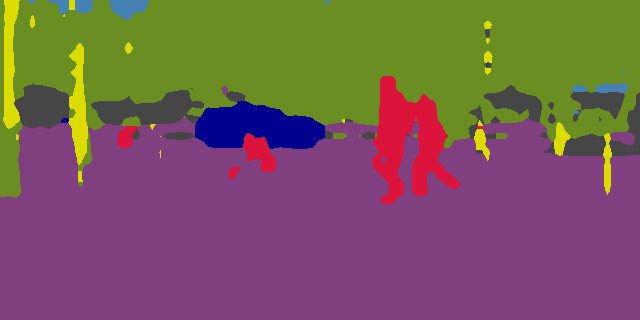} & \includegraphics[width=2.2cm]{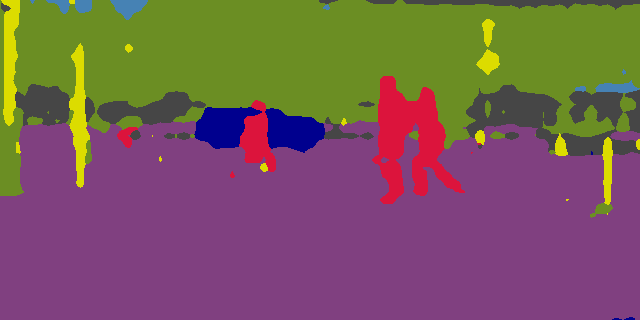}\\
        \cmidrule{2-8}
        
       \multirow{5}{*}{\rotatebox[origin=c]{90}{\parbox[c]{4cm}{\centering {\footnotesize IDD}}}} & \includegraphics[width=2.2cm,height=1.1cm]{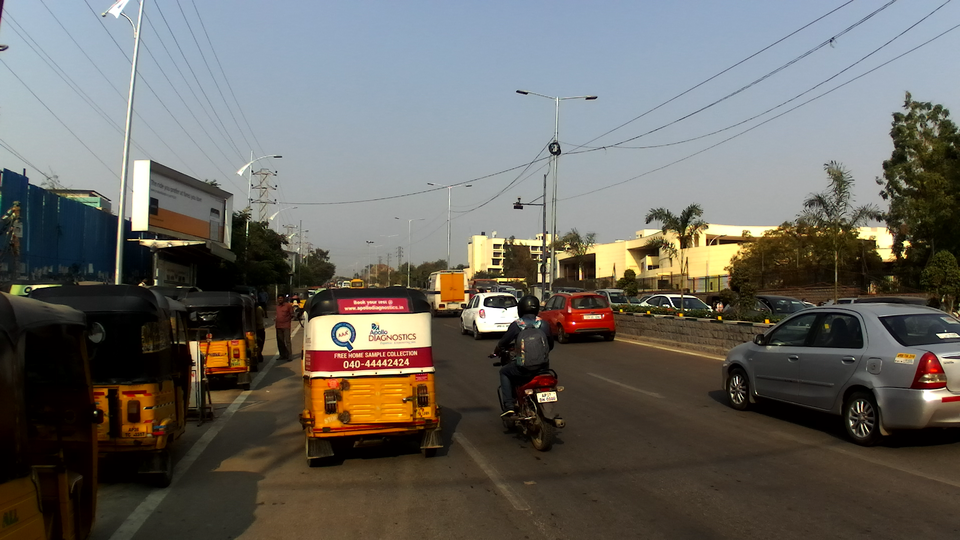} & \includegraphics[width=2.2cm]{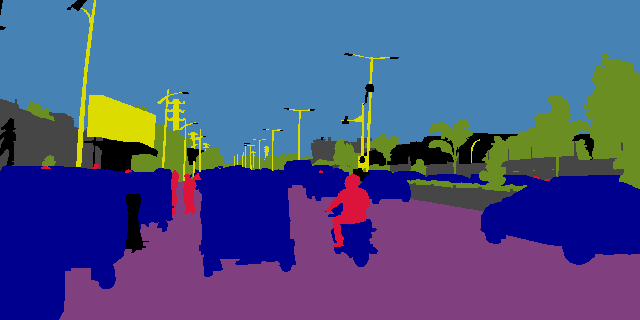} &
        \includegraphics[width=2.2cm]{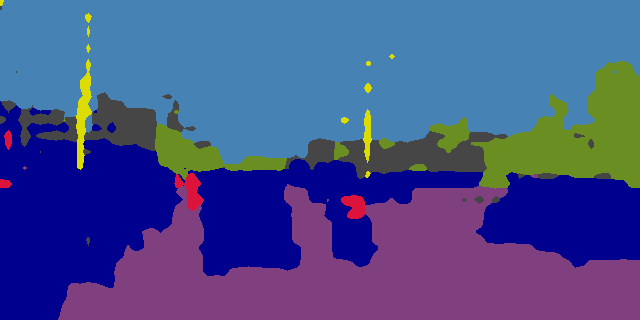}&
        \includegraphics[width=2.2cm]{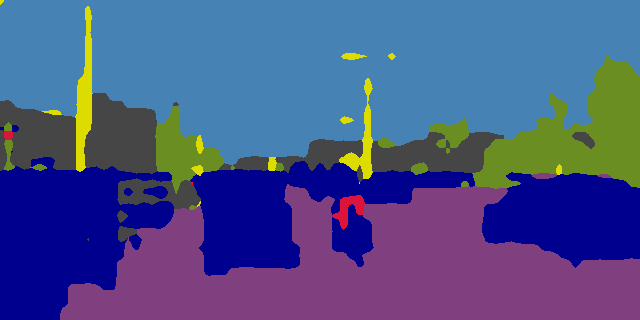} &\includegraphics[width=2.2cm]{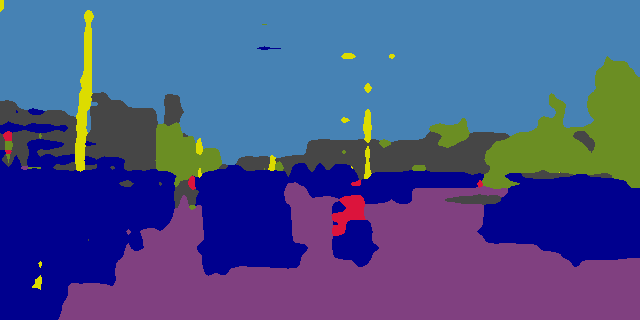} &\includegraphics[width=2.2cm]{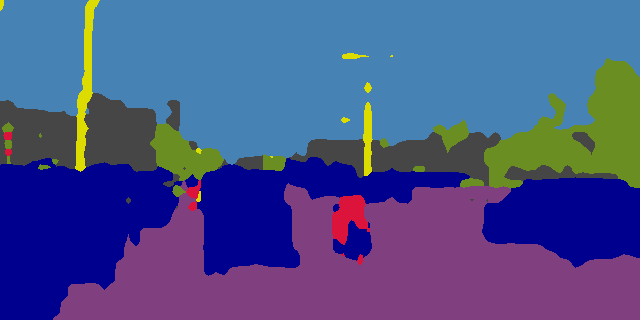} & \includegraphics[width=2.2cm]{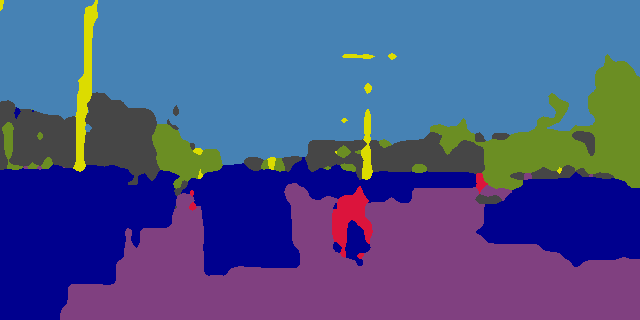}\\
       & \includegraphics[width=2.2cm,height=1.1cm]{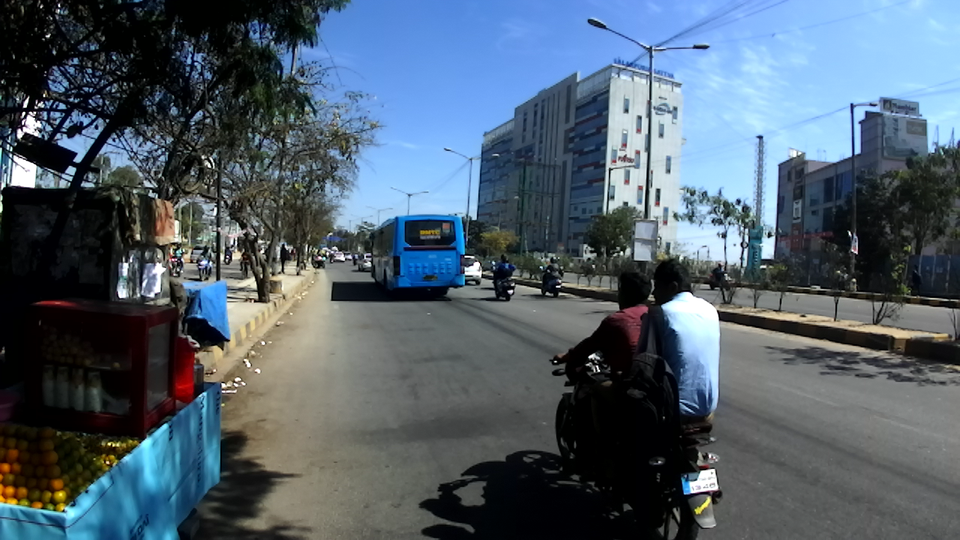} & \includegraphics[width=2.2cm]{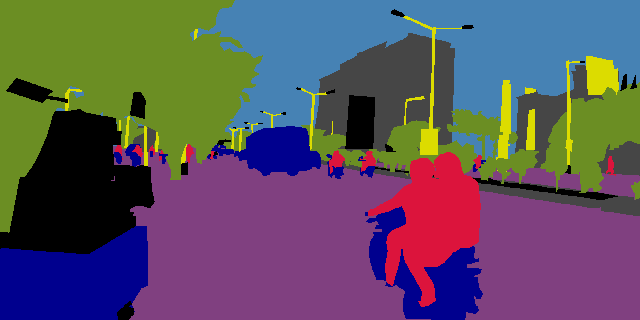} &\includegraphics[width=2.2cm]{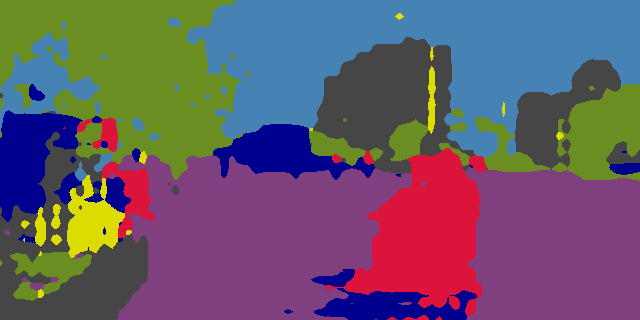}&
        \includegraphics[width=2.2cm]{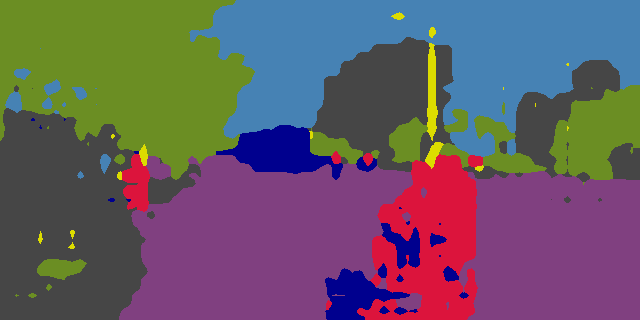}  &\includegraphics[width=2.2cm]{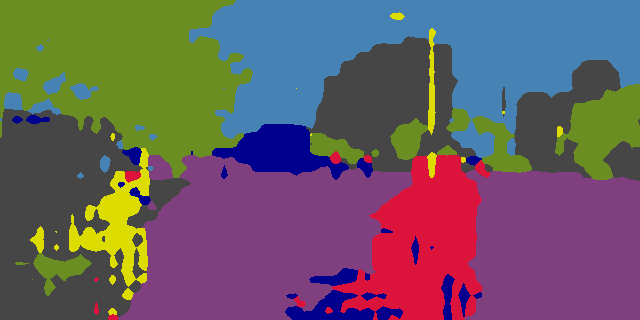}& \includegraphics[width=2.2cm]{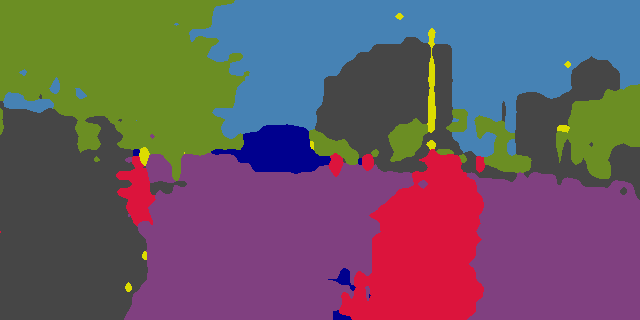} & \includegraphics[width=2.2cm]{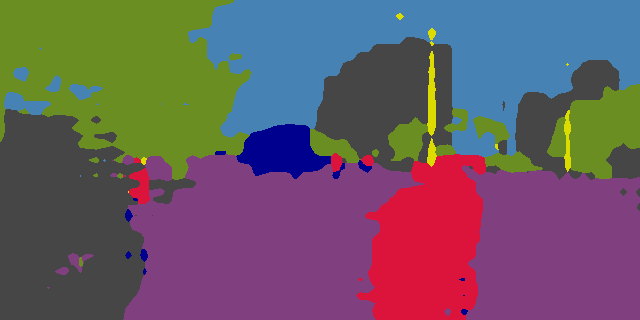}\\
        & \includegraphics[width=2.2cm, height=1.1cm]{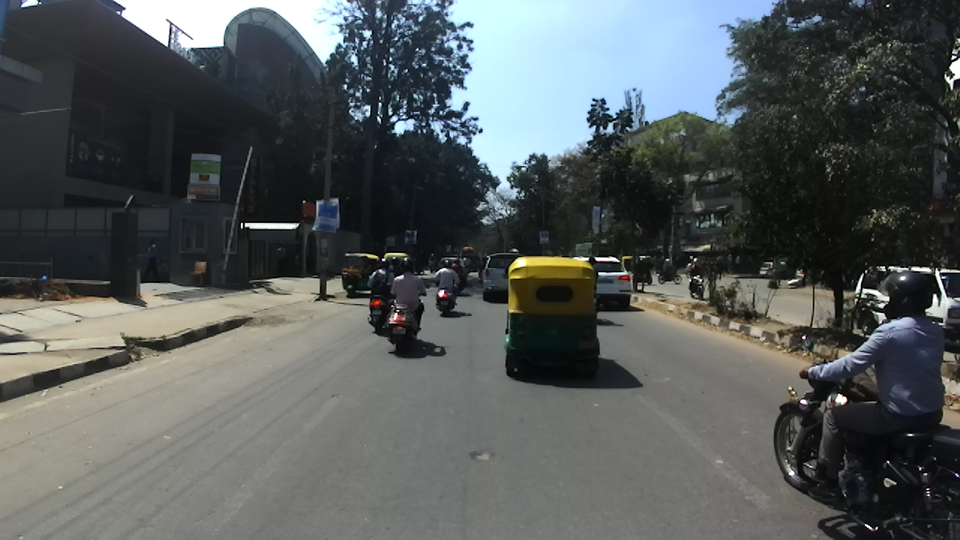} & \includegraphics[width=2.2cm]{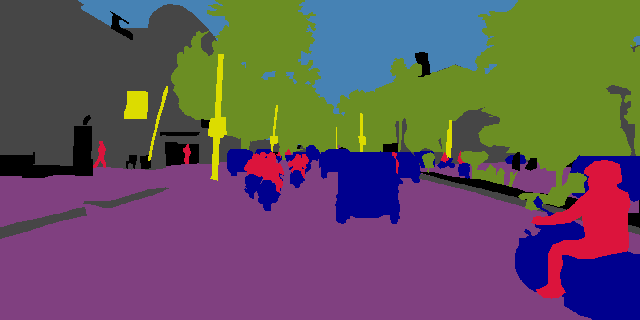} &\includegraphics[width=2.2cm]{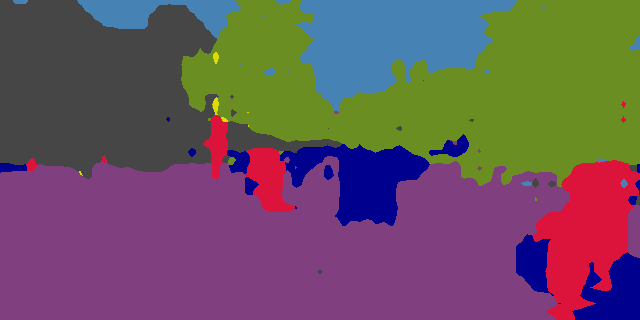}&
        \includegraphics[width=2.2cm]{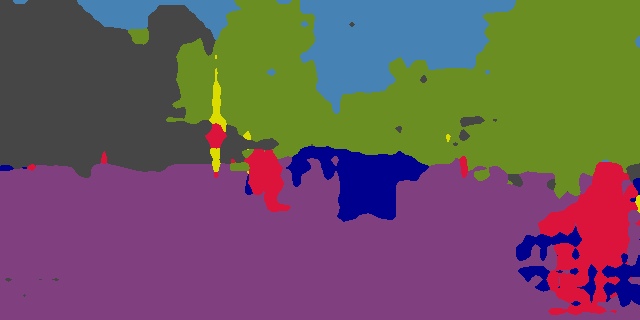}  &\includegraphics[width=2.2cm]{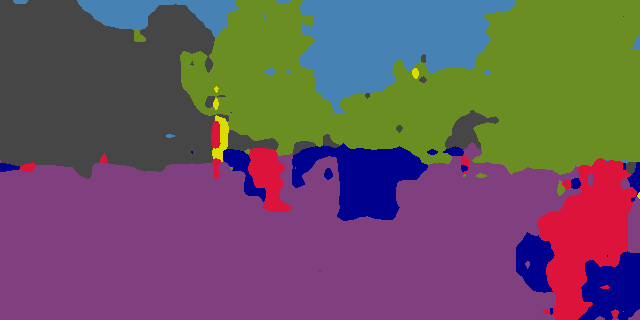}& \includegraphics[width=2.2cm]{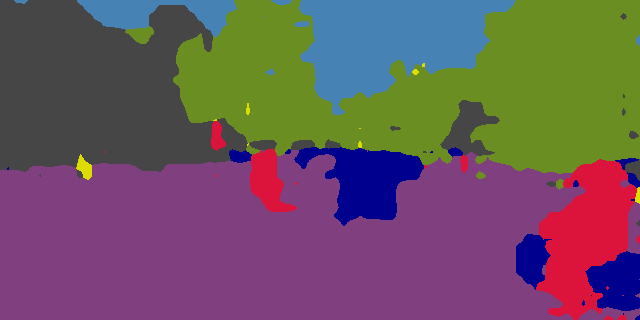} & \includegraphics[width=2.2cm]{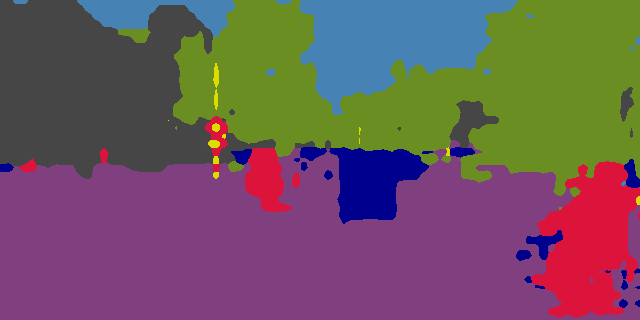}\\
         & \includegraphics[width=2.2cm, height=1.1cm]{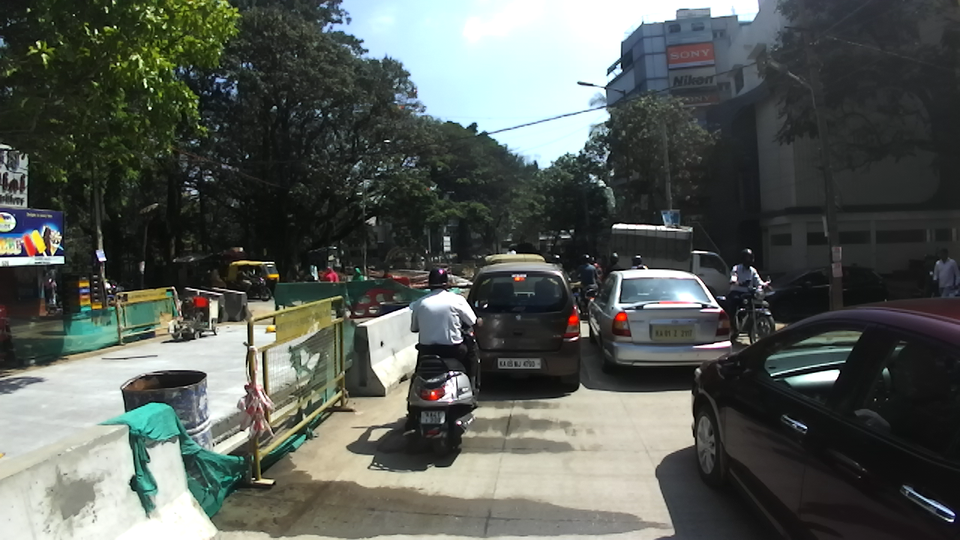} & \includegraphics[width=2.2cm]{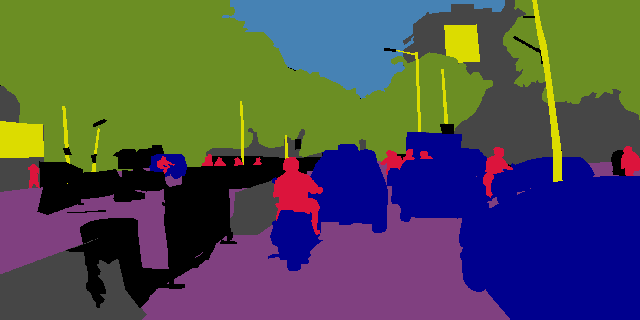} &\includegraphics[width=2.2cm]{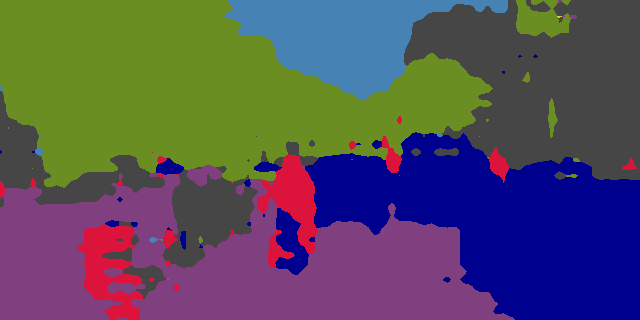}&
        \includegraphics[width=2.2cm]{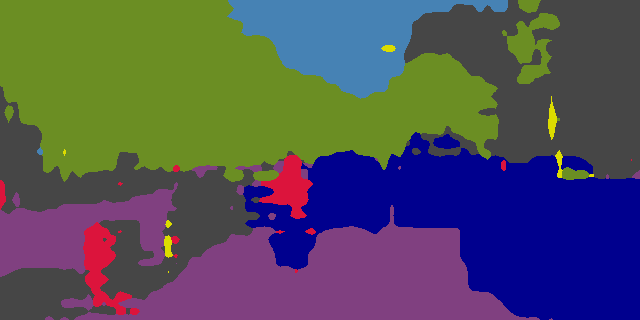}  &\includegraphics[width=2.2cm]{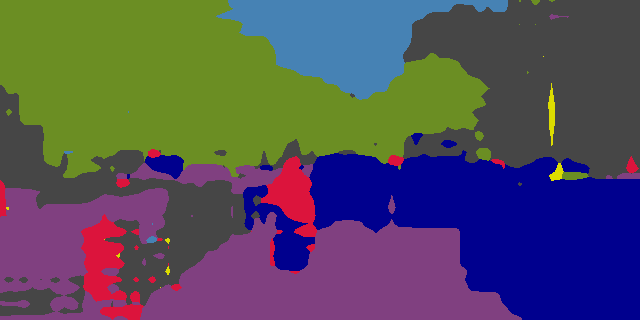}& \includegraphics[width=2.2cm]{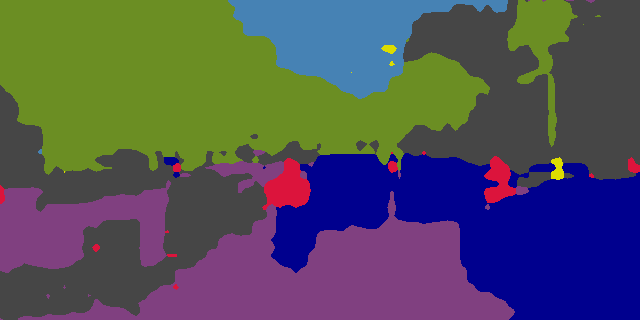} & \includegraphics[width=2.2cm]{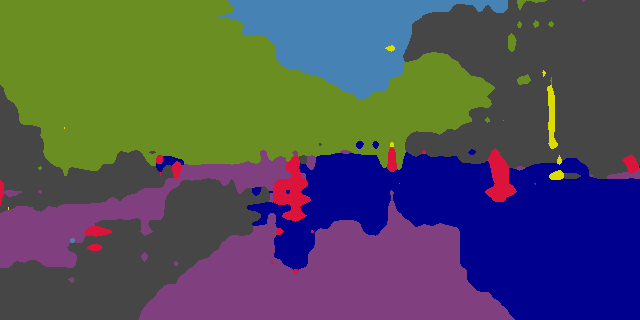}\\
         & \includegraphics[width=2.2cm, height=1.1cm]{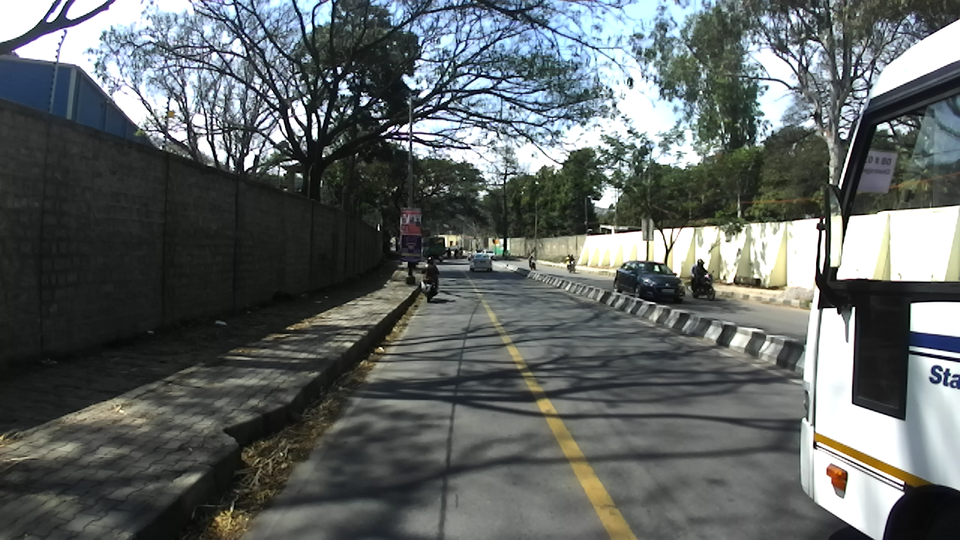} & \includegraphics[width=2.2cm]{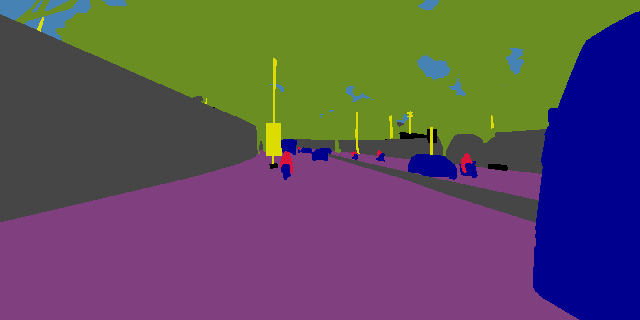} &\includegraphics[width=2.2cm]{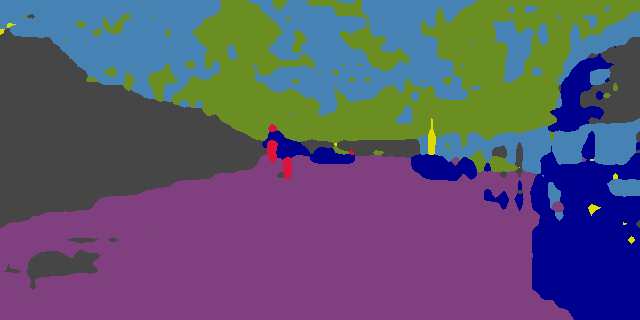}& \includegraphics[width=2.2cm]{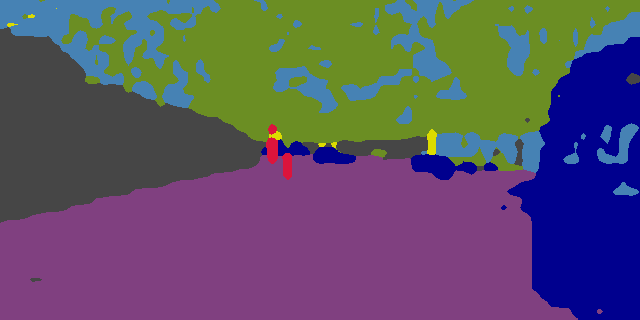}  &\includegraphics[width=2.2cm]{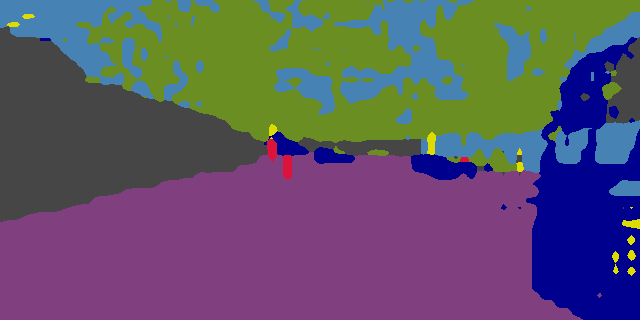}& \includegraphics[width=2.2cm]{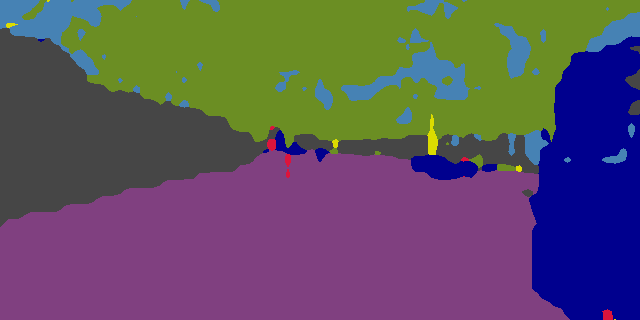} & \includegraphics[width=2.2cm]{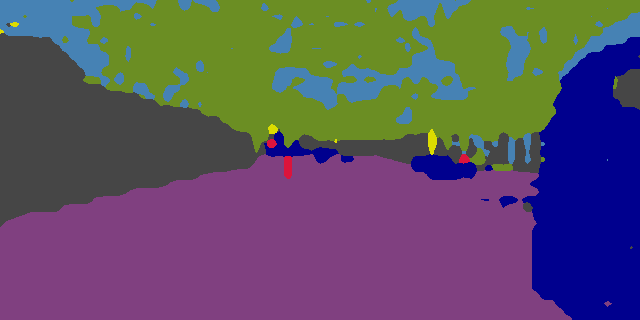}\\
    \end{tabular}}
    \vspace{-0.3cm}
    \caption{\textbf{Qualitative results in the GTA5\,$\shortrightarrow$\,Cityscapes\,$+$\,IDD setup.} (a) Test images from Cityscapes and IDD; (b) Ground-truth segmentation maps; Results of (c) single-target baseline trained on Cityscapes target, (d) single-target baseline trained on IDD target, (e) multi-target baseline, (f) proposed Multi-Dis. and (g) proposed MTKT. Both proposed multi-target frameworks give overall cleaner segmentation maps compared to the baselines. 
    }
    \label{fig:qualitative}
    \vspace{-0.4cm}
\end{figure*}

We visualize some qualitative results in Figure~\ref{fig:qualitative}. 

\begin{table}[t!]
\setlength\heavyrulewidth{0.25ex}
\aboverulesep=0ex
\belowrulesep=0ex
\resizebox{.48\textwidth}{!}{%
        \begin{tabular}{l| l | c| c c c c c c c|l|l<{\kern-\tabcolsep}}
            \toprule
            \multicolumn{12}{c}{\textbf{GTA5\,$\shortrightarrow$\,Cityscapes\,$+$\,Mapillary\,$+$\,IDD}}\\
            \midrule
            Method & Target & Train & \rotatebox{90}{flat} & \rotatebox{90}{constr.} & \rotatebox{90}{object} & \rotatebox{90}{nature} & \rotatebox{90}{sky} & \rotatebox{90}{human} & \rotatebox{90}{vehicle\,} & mIoU&\shortstack[l]{mIoU\\Avg.}\\
            \midrule
            \multirow{9}{*}{\shortstack[l]{Single-Target\\Baselines~\cite{vu-cvpr2019}}} & Cityscapes & \checkmark & 93.5 &	80.5 & 26.0 & 78.5 & 78.5 & 55.1 & 76.4 & 69.8 (*) & \multirow{3}{*}{65.5}  \\
            & Mapillary & - & 86.8 &	69.0 &	30.2 &	71.2 &	91.5 &	35.3 &	59.5 &	63.3$_{\down{\,6.3}}$ & \\
            & IDD & - & 91.3 &	52.3 &	13.3 &	76.1 &	88.7 &	46.7 &	74.8 &	63.3$_{\down{\,1.8}}$ & \\
            \cmidrule(lr){2-12}
            & Cityscapes & - & 89.3 &	79.3 &	19.5 &	76.9 &	84.6 &	47.7 &	63.0 &	65.8$_{\down{\,4.0}}$ & \multirow{3}{*}{66.7}  \\ 
            & Mapillary & \checkmark & 89.5 &	72.6 &	31.0 &	75.3 &	94.1 &	50.7 &	73.8 &	69.6 (*) & \\
            & IDD & - & 91.7 &	54.3 &	13.0 &	77.3 &	92.3 &	47.4 &	76.8 &	64.7$_{\down{\,0.4}}$ & \\
            \cmidrule(lr){2-12}
            & Cityscapes & - & 78.6 &	79.2 &	24.8 &	77.6 &	83.6 &	48.7 &	44.8 &	62.5$_{\down{\,7.3}}$ & \multirow{3}{*}{65.5}  \\
            & Mapillary & - & 88.5 & 71.2 &	32.4 & 72.8 & 92.8 & 51.3 &	73.7 & 69.0$_{\down{\,0.6}}$ & \\
            & IDD & \checkmark & 91.2 &	53.1 &	16.0 &	78.2 &	90.7 &	47.9 &	78.9 &	65.1 (*) & \\
            \midrule
            \multirow{3}{*}{\shortstack[l]{Multi-Target\\Baseline~\cite{vu-cvpr2019}}} & Cityscapes & \checkmark & 93.6 &	80.6 &	26.4 &	78.1 &	81.5 &	51.9 &	76.4 &	69.8$_-$ & \multirow{3}{*}{67.8}  \\
            & Mapillary & \checkmark & 89.2 &	72.4 &	32.4 &	73.0 &	92.7 &	41.6 &	74.9 &	68.0$_{\down{\,1.6}}$ & \\
            & IDD & \checkmark & 92.0 &	54.6 &	15.7 &	77.2 &	90.5 &	50.8 &	78.6 &	65.6$_{\up{\,0.5}}$ & \\
            \midrule
            \rowcolor[gray]{.92} & Cityscapes & \checkmark & 94.6 &	80.0 &	20.6 &	79.3 &	84.1 &	44.6 &	78.2 &	68.8$_{\down{\,1.0}}$ & \\
            \rowcolor[gray]{.92}& Mapillary & \checkmark & 89.0 &	72.5 &	29.3 &	75.5 &	94.7 &	50.3 &	78.9 &	70.0$_{\up{\,0.4}}$ & \\
            \rowcolor[gray]{.92} \multirow{-3}{*}{Multi-Dis.}& IDD & \checkmark & 91.6 &	54.2 &	13.1 &	78.4 &	93.1 &	49.6 &	80.3 &	65.8$_{\up{\,0.7}}$ & \multirow{-3}{*}{68.2} \\
            \midrule
            \rowcolor[gray]{.92} & Cityscapes & \checkmark & 94.6 &	80.7 &	23.8 &	79.0 &	84.5 &	51.0 &	79.2 &	\textbf{70.4}$_{\up{\,\mathbf{0.6}}}$ &  \\ 
            \rowcolor[gray]{.92}& Mapillary & \checkmark & 90.5 &	73.7 &	32.5 &	75.5 &	94.3 &	51.2 &	80.2 &	\textbf{71.1}$_{\up{\,\mathbf{1.5}}}$ & \\
            \rowcolor[gray]{.92}\multirow{-3}{*}{MTKT}& IDD & \checkmark & 91.7 &	55.6 &	14.5 &	78.0 &	92.6 &	49.8 &	79.4 &	\textbf{65.9}$_{\up{\,\mathbf{0.8}}}$ &  \multirow{-3}{*}{\textbf{69.1}} \\
           
            \bottomrule
        \end{tabular}
    }
\caption{
\textbf{Results on GTA5\,$\shortrightarrow$\,Cityscapes\,$+$\,Mapillary\,$+$\,IDD ($T=3$).}
Organization as in Tab.\,\ref{tab:G-CM}.
}
\label{tab:G-CMI}
\vspace{-0.3cm}
\end{table}

\begin{table}[t!]
\setlength\heavyrulewidth{0.25ex}
\aboverulesep=0ex
\belowrulesep=0ex
\resizebox{.48\textwidth}{!}{%
        \begin{tabular}{l| l | c | c c c c c c c|l|l<{\kern-\tabcolsep}}
            \toprule
            \multicolumn{12}{c}{\textbf{Cityscapes $\shortrightarrow$ Mapillary\,$+$\,IDD}}\\
            \midrule
            Method & Target & Train & \rotatebox{90}{flat} & \rotatebox{90}{constr.} & \rotatebox{90}{object} & \rotatebox{90}{nature} & \rotatebox{90}{sky} & \rotatebox{90}{human} & \rotatebox{90}{vehicle\,} & mIoU & \shortstack[l]{mIoU \\ Avg.} \\
            \midrule
            \multirow{4}{*}{\shortstack[l]{Single-Target\\Baselines~\cite{vu-cvpr2019}}} & Mapillary & \checkmark & 87.4 & 65.9 & 28.2 & 72.8 & 92.1 & 46.9 & 72.7 & 66.6 (*)&\multirow{2}{*}{65.8}\\
            & IDD & - & 91.8 & 52.2 & 15.9 & 80.2 & 91.1 & 45.7 & 77.6 & 65.0$_{\down{\,2.3}}$ \\
            \cmidrule(lr){2-12}
            & Mapillary & - & 88.2 & 70.0 & 28.5 & 75.4 & 93.6 & 49.1 & 76.7 & 68.8$_{\up{\,2.2}}$ &\multirow{2}{*}{68.0}\\
            & IDD & \checkmark & 93.2 & 53.4 & 16.5 & 83.4 & 93.4 & 51.4 & 79.5 & 67.3 (*)\\
            \midrule
            \multirow{2}{*}{\shortstack[l]{Multi-Target\\Baseline~\cite{vu-cvpr2019}}} & Mapillary  & \checkmark & 87.7 & 65.9 & 29.0 & 73.2 & 91.5 & 47.9 & 75.7 & 67.3$_{\up{\,0.7}}$&\multirow{2}{*}{67.0}\\
            & IDD  & \checkmark & 93.3 & 53.0 & 17.2 & 82.8 & 92.2 & 49.3 & 79.6 & 66.8$_{\down{\,0.5}}$ \\
            \midrule
            \rowcolor[gray]{.92} & Mapillary & \checkmark & 88.6 & 70.9 & 29.6 & 75.8 & 94.7 & 49.2 & 76.1 & 69.3$_{\up{\,2.7}}$ &\\
            \rowcolor[gray]{.92} \multirow{-2}{*}{Multi-Dis.}& IDD & \checkmark &  92.8 & 52.8 & 17.0 & 83.1 & 94.2 & 48.5 & 77.4 & 66.5$_{\down{\,0.8}}$ & \multirow{-2}{*}{67.9} \\
            \midrule
            \rowcolor[gray]{.92} & Mapillary & \checkmark & 88.3 & 70.4 & 31.6 & 75.9 & 94.4 & 50.9 & 77.0 & \textbf{69.8}$_{\up{\,\mathbf{3.2}}}$ &\\
            \rowcolor[gray]{.92}\multirow{-2}{*}{MTKT}& IDD & \checkmark & 93.6 & 54.9 & 18.6 & 84.0 & 94.5 & 53.4 & 79.2 & \textbf{68.3}$_{\up{\,\mathbf{1.0}}}$ & \multirow{-2}{*}{\textbf{69.0}} \\
            \bottomrule
        \end{tabular}
    }
\caption{
\textbf{Results of city-2-city multi-target UDA on Cityscapes\,$\shortrightarrow$\,Mapillary\,$+$\,IDD.} Organization as in Tab.\,\ref{tab:G-CM}.
}
\label{tab:C-MI}
\vspace{-0.3cm}
\end{table}

\smallskip\noindent{\textbf{GTA5\,$\shortrightarrow$\,Cityscapes\,$+$\,Mapillary\,$+$\,IDD.}~}
We consider a more challenging setup involving three target domains -- Cityscapes, Mapillary and IDD -- and show results in Table\,\ref{tab:G-CMI}. 
With more target domains, the same conclusions hold.
In terms of average mIoU, the multi-discriminator model marginally improves over the multi-target baseline.
The MTKT model significantly outperforms all other models with $69.1\%$ mIoU\,Avg.
Moreover, when compared to the per-target baselines, MTKT is the only model to show improvement on every target domain.

\smallskip\noindent{\textbf{Cityscapes $\shortrightarrow$ Mapillary\,$+$\,IDD.}~}
Finally, we experiment on a realistic city-2-city setup with Cityscapes as source and Mapillary and IDD as target domains.
The results are shown in Table~\ref{tab:C-MI}.
Interestingly, on Mapillary, the single-target model trained on IDD achieves better results than the one trained only on Mapillary.
We conjecture that the domain gap between Cityscapes and Mapillary is less than the one between Cityscapes and IDD; The extra data diversity coming from IDD improves the single-target IDD-only model generalization and helps mitigate the small Cityscapes-Mapillary domain gap.
Another observation is that the IDD-only model outperforms the multi-target baseline.
This indicates the disadvantage of the naive dataset merging strategy: Not only complementary signals but also conflicting/negative ones get transferred.
The two proposed models outperform the multi-target baseline; MTKT obtains the best performance overall.
Again in this realistic setup, we showcase the advantages of our methods, especially the multi-target knowledge transfer model.

\smallskip\noindent{\textbf{Conclusions.}~}
These four sets of experiments demonstrate that the proposed multi-target frameworks consistently deliver competitive performance on the multiple target domains they are trained for. MTKT always gives the best performance, both in per-target and average mIoUs, compared to the baselines and to the multi-discriminator model. Note that our models are compatible with techniques such as 
image translation~\cite{hoffman2017cycada,wu2018dcan,yang2020fda} or pseudo-labeling self-training~\cite{li-cvpr2019,saporta-cvpr2020, zou2018unsupervised}, from which they could benefit. In particular, we show next with additional experiments how to use pseudo-labeling~\cite{saporta-cvpr2020} with MTKT.

\subsection{Further Experiments}
\label{sec:further_exp}
\begin{table}[t]
\setlength\heavyrulewidth{0.25ex}
\aboverulesep=0ex
\belowrulesep=0ex
\setlength\extrarowheight{4pt}
\resizebox{.48\textwidth}{!}{%
       \begin{tabular}{r | c c | c c c c }
            \toprule
            \multicolumn{7}{c}{\textbf{GTA5\,$\shortrightarrow$\,Cityscapes\,$+$\,IDD}}\\
            \midrule
            Method  & M-T base.  & \shortstack{\ \\M-T base.\\+ PL}  & MTKT  & \shortstack{MTKT\\+ PL (1)} & \shortstack{MTKT\\+ PL (2)} & \shortstack{MTKT\\+ PL (3)} \\
            \midrule
            mIoU Avg. & 67.4 & 68.9 & 68.2 & 69.8 & 69.7 & \textbf{69.9} \\
            \bottomrule
        \end{tabular}
        }%
\caption{
\textbf{Additional impact of pseudo-labeling (PL).} Trained models are refined with one step of 
ESL~\cite{saporta-cvpr2020} (pseudo-labeling with predictive entropy as selection criteria). For MTKT, pseudo-labels are extracted for each target domain with the associated teacher head, and used either (1) to refine this head only, (2) to refine this head and to back-propagate KL-loss only on the pixels with predictions compliant with pseudo-labels or (3) to refine both this head and the target-agnostic model.
}
\label{tab:ESL}
\vspace{-0.2cm}
\end{table}

\smallskip\noindent{\textbf{Additional Impact of Pseudo-Labeling.~}}
Pseudo-labeling (PL) is a strategy that has become quite popular in UDA for semantic segmentation~\cite{li-cvpr2019,saporta-cvpr2020, zou2018unsupervised}. 
It can be easily combined with our multi-target frameworks. Taking for instance the recently-proposed ESL~\cite{saporta-cvpr2020},
we consider three ways to adapt 
its pseudo-labeling strategy to the MTKT architecture. In all of them, we collect pseudo-labels in each target domain using the corresponding target-specific classifier and use them as additional self-supervision for these target-specific heads; In the second method we also use these pseudo-labels to restrict the back-propagation of the KL losses to pixels that are correctly classified according to these pseudo-labels; In the third method, they are also used to refine the target-agnostic classifier.
We report in Table~\ref{tab:ESL} the results of the models trained with these three PL-based refinement strategies on GTA5\,$\shortrightarrow$\,Cityscapes\,$+$\,IDD and compare them to the baseline trained with ESL. The three ways of extending MTKT with PL result in similar performance gains of  at least $+1.6\%$ mIoU Avg. This demonstrates 
that knowledge transfer is complementary to pseudo-labeling. 
Moreover, MTKT with ESL outperforms the baseline with ESL by $+1.7\%$ mIoU Avg.

\begin{table}[t]
\setlength\heavyrulewidth{0.25ex}
\aboverulesep=0ex
\belowrulesep=0ex
\resizebox{.48\textwidth}{!}{%
        \begin{tabular}{l | l| l | c c c c c c c|c}
            \toprule
            \rotatebox{90}{setup} & Method & Test set & \rotatebox{90}{flat} & \rotatebox{90}{constr.} & \rotatebox{90}{object} & \rotatebox{90}{nature} & \rotatebox{90}{sky} & \rotatebox{90}{human} & \rotatebox{90}{vehicle} & mIoU \\

            \midrule
            \multirow{3}{*}{\rotatebox{90}{\scriptsize\textbf{G$\shortrightarrow$ C + I}}} & M-T Baseline & \multirow{3}{*}{Mapillary} & 88.4 & 71.0 & \textbf{31.0} & 72.4 & 92.0 & 37.4 & 74.7 & 66.7\\
            & Multi-Dis. &  & 89.2 & 72.1 & 21.7 & 73.8 & \textbf{94.0} & 34.8 & 75.9 & 65.9 \\

            & MTKT &  & \textbf{89.8} & \textbf{74.0} & 30.4 & \textbf{74.1} & 93.6 & \textbf{52.6} & \textbf{79.4} & \textbf{70.6} \\
            
            \midrule
            \multirow{3}{*}{\rotatebox{90}{\scriptsize\textbf{G$\shortrightarrow$ C + M\,}}} & M-T Baseline & \multirow{3}{*}{IDD} & \textbf{91.6} & 54.7 & \textbf{13.9} & 76.5 & 90.9 & 48.3 & 77.5 & 64.8\\

            & Multi-Dis. &  & 91.2  & 54.6 & 12.9 & \textbf{77.7} & \textbf{92.5} & 50.3 & 78.6& \textbf{65.4}\\

            & MTKT &  & 91.5 & \textbf{56.1} & 12.3 & 76.1 & 90.9 & \textbf{51.4} & \textbf{79.2} & \textbf{65.4}\\
            \bottomrule
        \end{tabular}
    }
\caption{\textbf{Direct transfer to new target.} Multi-target models are tested on a new unseen domain: (Top) 
GTA5\,$\shortrightarrow$\,Cityscapes\,$+$\,IDD, tested on Mapillary; (Bottom) 
GTA5\,$\shortrightarrow$\,Cityscapes\,$+$\,Mapillary, tested on IDD.}
\label{tab:DT}
\vspace{-0.3cm}
\end{table}

\smallskip\noindent{\textbf{Direct Transfer to a New Dataset.}~}
We consider a direct transfer setup in which the models see no images
from the test domain during training: This experiment highlights how well the models can generalize to new previously-unseen domains.
We report in Table~\ref{tab:DT} the results of such a direct transfer to a new dataset in different setups. The models are trained on GTA5\,$\shortrightarrow$\,Cityscapes\,$+$\,IDD (resp. on GTA5\,$\shortrightarrow$\,Cityscapes\,$+$\,Mapillary) and tested on Mapillary (resp. IDD). 
On both setups, MTKT shows better performance in terms of mIoU compared to the baselines on the new domain. In the first one in particular, with Mapillary as the new test domain, 
MTKT outperforms the multi-target baseline by $+3.9\%$. What is particularly noticeable in this setup is the performance on the \emph{human} class: While we observe an IoU of around $50\%$ in the main results on domain adaptation to Mapillary (e.g. in Tab.\,\ref{tab:G-CM}), the direct transfer results of the multi-target baseline and of Multi-Dis. drop under $38\%$ on this class; Differently, MTKT manages to get similar performance with $52.8\%$ IoU on \emph{human}. This experiment hints at the ability of MTKT to better generalize to new unseen domains. 

\section{Conclusion}
This work addresses the new problem of unsupervised adaptation to multiple target domains in semantic segmentation.
We discuss the challenges that this UDA setup raises in terms of distribution alignment and of joint learning.
That leads to two novel frameworks: The multi-discriminator approach extends single-target UDA to handle pair-wise domain alignment; The multi-target knowledge transfer approach alleviates the instability of multi-domain adversarial learning with a multi-teacher/single-student distillation mechanism.
In the context of driving scenes, we propose four experimental setups, varying the type of source-target gaps and the number of target domains.
Our approaches outperform all baselines on these four setups, which are representative of real-world applications.
Further experiments additionally show that our frameworks can be combined to state-of-the-art pseudo-labeling strategies and that the proposed learning schemes help to generalize to previously-unseen datasets.
This work thus contributes to the recent research line in domain adaptation toward more practical use cases.
With the same goal, future research directions may consider more complex mixes of source and target domains, making use of several labeled and unlabeled datasets.

\clearpage
\paragraph{Acknowledgments} This work was granted access to the HPC resources of IDRIS under the allocation 2020-AD011012153 made by GENCI.
\appendix
\section{Mapping Classes to Super Classes}
Tabs.\,\ref{tab:cm_cityscapes}, \ref{tab:cm_gtav}, \ref{tab:cm_mapillary} and \ref{tab:cm_idd} present how the original classes in the 4 considered datasets are mapped to 7 shared super classes.

\begin{table}[!h]
    \setlength\heavyrulewidth{0.25ex}
    \aboverulesep=0ex
    \belowrulesep=0ex
    \centering
    \small
    \resizebox{.38\textwidth}{!}{
    \begin{tabular}{l|c|c|c}
        \toprule
        Name & Orig. Id & Used? & Label \\
        \midrule
        unlabeled & 0 & & void\\
        ego vehicle & 1 & & void\\
        rectification border & 2 & & void\\
        out of roi & 3 & & void\\
        static & 4 & & void\\
        dynamic & 5 & & void\\
        ground & 6 & & void\\
        road & 7 & \checkmark & \Flat\\
        sidewalk & 8 & \checkmark & \Flat\\
        parking & 9 & & \Flat\\
        rail track & 10 & & \Flat\\
        building & 11 & \checkmark & \construction\\
        wall & 12 & \checkmark &\construction\\
        fence & 13 & \checkmark &\construction\\
        guard rail & 14 & & \construction\\
        bridge & 15 & & \construction\\
        tunnel & 16 & & \construction\\
        pole & 17 & \checkmark & \object\\
        polegroup & 18 & & \object\\
        traffic light & 19 & \checkmark &\object\\
        traffic sign & 20 & \checkmark &\object\\
        vegetation & 21 & \checkmark &\nature\\
        terrain & 22 & \checkmark &\nature\\
        sky & 23 & \checkmark &\sky\\
        person & 24 & \checkmark &\human\\
        rider & 25 & \checkmark &\human\\
        car & 26 & \checkmark & \vehicle\\
        truck & 27 & \checkmark &\vehicle\\
        bus & 28 &\checkmark &\vehicle\\
        caravan & 29 & & \vehicle\\
        trailer & 30 & & \vehicle\\
        train & 31 & \checkmark &\vehicle\\
        motorcycle & 32 & \checkmark &\vehicle\\
        bicycle & 33 & \checkmark &\vehicle\\
        license plate & -1 & & \vehicle\\
        \bottomrule
    \end{tabular}}
    \caption{\textbf{Mapping classes of Cityscapes.} 
    First two columns: names and ids of original classes;  Third column: indication of the use or not of a class during training and test; Last column: Super classes that are mapped to.
    }
    \label{tab:cm_cityscapes}
\end{table}

\begin{table}[!t]
    \setlength\heavyrulewidth{0.25ex}
    \aboverulesep=0ex
    \belowrulesep=0ex
    \centering
    \small
    \resizebox{.32\textwidth}{!}{
    \begin{tabular}{l|c|c|c}
        \toprule
        Name & Orig. Id & Used? & Label \\
        \midrule
        road & 0 & \checkmark & \Flat\\
        sidewalk & 1 & \checkmark & \Flat\\
        building & 2 & \checkmark & \construction\\
        wall & 3 & \checkmark &\construction\\
        fence & 4 & \checkmark &\construction\\
        pole & 5 & \checkmark & \object\\
        traffic light & 6 & \checkmark & \object\\
        traffic sign & 7 & \checkmark & \object\\
        vegetation & 8 & \checkmark & \nature\\
        terrain & 9 & \checkmark & \nature\\
        sky & 10 & \checkmark & \sky\\
        person & 11 & \checkmark & \human\\
        rider & 12 & \checkmark & \human\\
        car & 13 & \checkmark & \vehicle\\
        truck & 14 & \checkmark & \vehicle\\
        bus & 15 & \checkmark & \vehicle\\
        train & 16 & \checkmark & \vehicle\\
        motorcycle & 17 & \checkmark & \vehicle\\
        bicycle & 18 & \checkmark & \vehicle\\
        unlabeled & -1 & & void\\
        \bottomrule
    \end{tabular}}
    \caption{\textbf{Mapping classes of GTA5.} Organization as in Table~\ref{tab:cm_cityscapes}. }
    \label{tab:cm_gtav}
\end{table}

\begin{table}[!t]
    \setlength\heavyrulewidth{0.25ex}
    \aboverulesep=0ex
    \belowrulesep=0ex
    \centering
    \small
    \resizebox{.38\textwidth}{!}{
    \begin{tabular}{l|c|c|c}
        \toprule
        Name & Orig. Id & Used? & Label \\
        \midrule
        bird & 0 & & other\\
        ground animal & 1 & & other\\
        curb & 2 & \checkmark & \construction\\
        fence & 3 & \checkmark & \construction\\
        guard rail & 4 & \checkmark & \construction\\
        barrier & 5 & \checkmark & \construction\\
        wall & 6 & \checkmark & \construction\\
        bike lane & 7 & \checkmark & \Flat\\
        crosswalk - plain & 8 & \checkmark & \Flat\\
        curb cut & 9 & \checkmark & \Flat\\
        parking & 10 & \checkmark & \Flat\\
        pedestrian area & 11 & \checkmark & \Flat\\
        rail track & 12 & \checkmark & \Flat\\
        road & 13 & \checkmark & \Flat\\
        service lane & 14 & \checkmark & \Flat\\
        sidewalk & 15 & \checkmark & \Flat\\
        bridge & 16 & \checkmark & \construction\\
        building & 17 & \checkmark & \construction\\
        tunnel & 18 & \checkmark & \construction\\
        person & 19 & \checkmark & \human\\
        bicyclist & 20 & \checkmark & \human\\
        motorcyclist & 21 & \checkmark & \human\\
        other rider & 22 & \checkmark & \human\\
        lane marking - crosswalk & 23 & \checkmark & \Flat\\
        lane marking - general  & 24 & \checkmark & \Flat\\
        mountain & 25 & & \nature\\
        sand & 26 & & \nature\\
        sky & 27 & \checkmark & \sky\\
        snow & 28 & & \nature\\
        terrain & 29 & \checkmark & \Flat\\
        vegetation & 30 & \checkmark &\nature \\
        water & 31 & & \nature\\
        banner & 32 & & \object\\
        bench & 33 & & \object\\
        bike rack & 34 & & \object\\
        billboard & 35 & & \object\\
        catch basin & 36 & & \object\\
        cctv camera & 37 & & \object\\
        fire hydrant & 38 & & \object\\
        junction box & 39 & & \object\\
        mailbox & 40 & & \object\\
        manhole & 41 & & \object\\
        phone booth & 42 & & \object\\
        pothole & 43 & \checkmark & \object\\
        street light & 44 & \checkmark & \object\\
        pole & 45 & \checkmark & \object\\
        traffic sign frame & 46 & \checkmark & \object\\
        utility pole & 47 & \checkmark & \object\\
        traffic light & 48 & \checkmark & \object\\
        traffic sign (back) & 49 & \checkmark & \object\\
        traffic sign (front) & 50 & \checkmark & \object\\
        trash can & 51 & & \object\\
        bicycle & 52 & \checkmark & \vehicle\\
        boat & 53 & \checkmark & \vehicle\\
        bus & 54 & \checkmark & \vehicle\\
        car & 55 & \checkmark & \vehicle\\
        caravan & 56 & \checkmark & \vehicle\\
        motorcycle & 57 & \checkmark & \vehicle\\
        on rails & 58 & \checkmark & \vehicle\\
        other vehicle & 59 & \checkmark & \vehicle\\
        trailer & 60 & \checkmark & \vehicle\\
        truck & 61 & \checkmark & \vehicle\\
        wheeled slow & 62 & \checkmark & \vehicle\\
        car mount & 63 & & void\\
        ego vehicle & 64 & & void\\
        unlabeled & -1 & & void\\
        \bottomrule
    \end{tabular}
    }
    \caption{\textbf{Mapping classes of Mapillary Vistas.} Organization as in Table~\ref{tab:cm_cityscapes}. }
    \label{tab:cm_mapillary}
\end{table}

\begin{table}[!t]
\setlength\heavyrulewidth{0.25ex}
\aboverulesep=0ex
\belowrulesep=0ex
    \centering
    \small
    \resizebox{.38\textwidth}{!}{
    \begin{tabular}{l|c|c|c}
        \toprule
        Name & Orig. Id & Used? & Label \\
        \midrule
        road & 0 & \checkmark & \Flat\\
        parking & 1 & & \Flat \\
        drivable fallback & 2 & \checkmark & \Flat\\
        sidewalk & 3 & \checkmark & \Flat\\
        rail track & 4 & &  \Flat\\
        non-drivable fallback & 5 & & \Flat \\
        person & 6 & \checkmark & \human\\
        animal & 7 & & other\\
        rider & 8 & \checkmark &\human\\
        motorcycle & 9 & \checkmark &\vehicle\\
        bicycle & 10 & \checkmark &\vehicle\\
        autorickshaw & 11 & \checkmark &\vehicle\\
        car & 12 & \checkmark &\vehicle\\
        truck & 13 & \checkmark &\vehicle\\
        bus & 14 & \checkmark &\vehicle\\
        caravan & 15 & \checkmark &\vehicle\\
        trailer & 16 & \checkmark &\vehicle\\
        train & 17 & \checkmark &\vehicle\\
        vehicle fallback & 18 & \checkmark &\vehicle\\
        curb & 19 & \checkmark &\construction\\
        wall & 20 & \checkmark &\construction\\
        fence & 21 & \checkmark &\construction\\
        guard rail & 22 & & \construction\\
        billboard & 23 & \checkmark &\object\\
        traffic sign & 24 & \checkmark &\object\\
        traffic light & 25 & \checkmark &\object\\
        pole & 26 & \checkmark &\object\\
        polegroup & 27 & & \object\\
        obs-str-bar-fallback & 28 & &\object\\
        building & 29 & \checkmark &\construction\\
        bridge & 30 & & \construction\\
        tunnel & 31 & & \construction\\
        vegetation & 32 & \checkmark & \nature\\
        sky & 33 & \checkmark & \sky\\
        fallback background & 34 & & \object\\
        unlabeled & 35 & & void\\
        ego vehicle & 36 & & void\\
        rectification border & 37 & & void\\
        out of roi & 38 & & void\\
        license plate & 39 & & \vehicle\\
        \bottomrule
    \end{tabular}}
    \caption{\textbf{Mapping classes of IDD.} Organization as in Table~\ref{tab:cm_cityscapes}. }
    \label{tab:cm_idd}
\end{table}

\clearpage
{\small
\bibliographystyle{ieee_fullname}
\bibliography{egbib}
}

\end{document}